\documentclass[journal]{IEEEtran}
\usepackage{amssymb}
\usepackage{amsmath}
\usepackage{algorithmic}
\usepackage{algorithm}
\usepackage{array}
\usepackage{stfloats}
\usepackage{url}
\usepackage{verbatim}
\usepackage{graphicx}
\usepackage[numbers,sort&compress]{natbib}
\usepackage{newtxtext,newtxmath}
\usepackage{caption}
\usepackage{hyperref}
\hypersetup{
hidelinks,
pdftitle={Koopman Model Predictive Control of An Origami-Inspired Soft Exoskeleton for Knee Rehabilitation},
pdfauthor={Junxiang Wang, Han Zhang, Zehao Wang, Huaiyuan Chen, Pu Wang, Weidong Chen}
}
\usepackage{multirow}
\usepackage{makecell}
\usepackage{booktabs}
\usepackage{footnote}
\usepackage{amsfonts}
\usepackage{bm}
\usepackage{caption}
\usepackage{subcaption}

\DeclareMathOperator*{\diag}{diag}

\ifCLASSINFOpdf

\else

\fi

\begin{document}

\title{Koopman Model Predictive Control of An Origami-Inspired Soft Exoskeleton for Knee Rehabilitation}
\author{Junxiang~Wang$^{1,2}$, Han~Zhang$^{1,2}$, Zehao~Wang$^{1,2}$, Huaiyuan~Chen$^{1,2}$, Pu~Wang$^{3}$, and Weidong~Chen$^{1,2}$%
\thanks{This work was supported by the National Natural Science Foundation of China under Grant 62573286, the Natural Science Foundation of Shanghai under Grant 25ZR1401208, and the Shanghai Key Laboratory of Flexible Medical Robotics under Open Project SKLFMR-O213.}
\thanks{$^{1}$School of Automation and Intelligent Sensing, Institute of Medical Robotics, and Shanghai Key Laboratory of Flexible Medical Robotics, Tongren Hospital, Shanghai Jiao Tong University, Shanghai 200240, China.}
\thanks{$^{2}$Key Laboratory of System Control and Information Processing, Ministry of Education of China, Shanghai 200240, China.}
\thanks{$^{3}$The Seventh Affiliated Hospital, Sun Yat-sen University, Shenzhen 518107, China.}
\thanks{Han Zhang is the corresponding author (e-mail: zhanghan\_tc@sjtu.edu.cn; ORCID: 0000-0002-3905-0633).}}
\markboth{IEEE Transactions on Medical Robotics and Bionics}%
{Koopman Model Predictive Control of an Origami-Inspired Soft Exoskeleton for Knee Rehabilitation}

\maketitle

\begin{abstract}
Knee rehabilitation plays a critical role in restoring patients' mobility and functional independence. Traditional rigid rehabilitation exoskeletons are often bulky and cumbersome to wear, whereas soft pneumatic exoskeletons offer lightweight, wearable, and intrinsically compliant solutions that are better suited for human--robot interaction. However, achieving precise motion control for soft exoskeletons remains challenging due to the difficulty of accurately modeling pneumatic actuators and the patient-specific human--robot coupled dynamics during rehabilitation training.
To address these challenges, this paper proposes a Koopman-based Model Predictive Control (KMPC) framework for soft knee rehabilitation exoskeletons. The nonlinear human--robot coupled system is represented through a lifted linear Koopman model, enabling predictive control with explicit handling of constraints. In addition to actuation commands used to control valves and pumps, electromyography (EMG) signals are incorporated as system inputs, allowing the Koopman model to capture voluntary neuromuscular contribution and individual neuromuscular characteristics.
Experimental results on both healthy participants and patients demonstrate that the proposed framework improves model prediction accuracy and effectively captures subject-specific behaviors, thereby supporting EMG-informed subject-specific assistance within the tested seated knee-rehabilitation setting. Compared with conventional Proportional--Integral--Derivative (PID) control, the proposed KMPC approach achieves lower tracking errors and reduced actuation effort in both passive and active rehabilitation modes. Additional comparisons with Iterative Learning Control (ILC) further validate the tracking performance of the proposed controller.
\end{abstract}


\IEEEpeerreviewmaketitle

\section{Introduction}
Due to stroke and other neurological disorders, knee joint dysfunction significantly increases the risk of falling and hampers patients' functional recovery \cite{wang2025effect,lyu2019development}. The knee joint plays a central role in lower-limb function by supporting body weight and enabling smooth gait transitions, thereby facilitating essential daily activities such as walking and stair climbing. Consequently, the development of rehabilitation exoskeletons and their control strategies that enable precise and personalized knee assistance is critical for restoring patients' motor function. Traditional rigid exoskeletons have demonstrated effectiveness in providing strong support, accurate force transmission, and fast dynamic response \cite{9900071}. However, their limited intrinsic compliance and substantial weight pose fundamental challenges for human--robot interaction. In particular, the rigid structures make it difficult to achieve a precise alignment between the exoskeleton joint axis and the user's anatomical knee joint \cite{10057135}, which can degrade interaction performance \cite{8250095,10716507,9371725}. Such misalignment not only causes discomfort during wear, but may also interfere with patients' natural movement patterns during rehabilitation training. Moreover, the considerable weight of rigid exoskeletons often necessitates assistance from caregivers for donning and doffing \cite{rodriguez2021systematic}, thereby reducing usability and limiting the effective training duration. These factors collectively decrease patient compliance and may ultimately compromise rehabilitation outcomes \cite{CHEN201626}.

To overcome these limitations, soft exoskeletons have been developed as an alternative solution \cite{9460319,9116050,9785890}, employing actuation mechanisms such as Pneumatic Artificial Muscles (PAMs) \cite{6907562,ma2022design} and cable-driven systems \cite{9627581}. For seated knee rehabilitation scenarios, however, soft exoskeletons must strike a delicate balance between wearability and the ability to deliver reliable assistive torque over a range of motion consistent with knee kinematics. Cable-driven actuation does not generate joint torque directly, and the resulting shear forces can lead to anchor slippage and soft tissue deformation \cite{Thalman_Artemiadis_2020}. Meanwhile, conventional McKibben-type PAMs typically require high operating pressures to produce sufficient assistive torque \cite{9646244,9300132}, necessitating bulky pressure sources that hinder portability and home-based rehabilitation. These challenges highlight the need for knee rehabilitation soft exoskeletons that are both portable and capable of achieving high actuation efficiency.

In knee rehabilitation scenarios, soft exoskeletons are required to assist the motion of the lower limbs, which imposes substantial and time-varying loads on the actuation system. Achieving satisfactory dynamic performance under such conditions requires an accurate model of the soft actuators and an appropriate controller design. However, due to the inherently unstructured deformation of soft actuators during operation, obtaining precise analytical models remains particularly challenging. As a result, nonlinear control methods that rely on accurate system dynamics, such as backstepping or sliding mode control, are difficult to apply in practice. Hence carefully tuned Proportional--Integral--Derivative (PID) controllers are commonly adopted \cite{9016399}, yet their effectiveness often degrades across users due to inter-individual variability in human biomechanics. Adaptive control \cite{11142330,TOROOSSABA2024105774,11208847} and robust control \cite{7867797,11106485} have also been explored to address parametric uncertainties in the model, but these approaches typically require a nominal model as a starting point. When a soft exoskeleton is worn by a human during rehabilitation training, the system dynamics arise from the coupled interaction between the soft robot actuation and the human musculoskeletal system. The latter is highly personalized and difficult to characterize in clinical settings. Moreover, many adaptive and robust control methods are primarily formulated to compensate for model uncertainty, interaction variation, and external disturbances, whereas the subject's voluntary contribution in rehabilitation can provide informative input for modeling and control. This motivates a data-driven formulation that explicitly incorporates the human contribution into a control-oriented predictive model, rather than treating it only as an uncertainty or disturbance to be passively accommodated.

To address the above challenges, this work adopts the Koopman operator framework for modeling and control of soft knee rehabilitation exoskeletons. In particular, Koopman operator theory provides a principled approach to represent nonlinear dynamics through a lifted linear system in a high-dimensional space, thereby enabling the use of linear prediction and control tools for complex nonlinear systems \cite{Korda2018149,9277915}. Recent advances in deep learning-based Koopman methods \cite{8815339,9799788} further enable the learning of lifting functions and the corresponding linear dynamics directly from data. Since the collected data inherently reflects the actuator's nonlinearities, inter-individual variability, and human-robot coupling effects, these characteristics are naturally embedded in the learned lifted space. Moreover, to explicitly account for human contribution in the coupled dynamics, electromyography (EMG) signals are incorporated as additional inputs to the Koopman model. Since the EMG signals provide direct measurements of neuromuscular activation and hence the voluntary effort during rehabilitation training, this allows the resulting model to capture knee motion arising from both the muscle actuation and the soft exoskeleton assistance. 
Based on this idea, this paper develops a soft knee rehabilitation exoskeleton platform based on the origami-inspired pneumatic actuator in \cite{10361525}, and focuses on the modeling and control of the resulting human--robot coupled system. In particular, the chamber number, geometric dimensions, connector angle, and fixation structures are redesigned to meet the mechanical and ergonomic requirements of seated knee rehabilitation. Moreover, a subject-specific Deep Koopman model is learned by incorporating both the robot actuation input and EMG-derived human contribution, and a KMPC framework is then designed for real-time assistive control under input constraints. The proposed framework is evaluated on both healthy participants and stroke patients as an initial clinical validation, and the controller is benchmarked against both PID and Iterative Learning Control (ILC).

To summarize, the main contribution of this work is three-fold:
\begin{itemize}
    \item With the concept of delivering reliable assistive torque over a motion range aligned with natural knee kinematics while improving force transmission, alignment, and wearing stability, we re-design and optimize the origami-inspired pneumatic actuator in \cite{10361525} for knee rehabilitation, resulting in a portable and easy-to-wear soft knee rehabilitation exoskeleton.
    \item A human--robot coupled dynamic model is learned using a Deep Koopman Network, in which both EMG signals and actuation commands are incorporated as system inputs to enhance modeling accuracy.
    \item A KMPC framework is proposed for real-time control of the soft exoskeleton. Experimental results on both healthy participants and stroke patients demonstrate its ability to capture individual biomechanical characteristics and improve the tracking performance compared with PID and ILC in the tested rehabilitation setting. The data collection and training procedure is also practical under the current clinical experimental protocol.
\end{itemize}

The remainder of this paper is organized as follows. Section II presents the methodology, including the exoskeleton design, sensing and signal processing, safety mechanisms, the human--robot coupled Koopman modeling, and the KMPC formulation. Section III describes the experimental protocol, participants, experiment procedure, baselines, and evaluation metrics. Section IV reports the experimental results. Section V discusses the findings and limitations of the current study, including the scope of the present clinical validation. Finally, Section VI concludes the paper.

\section{Methodology}
This section introduces the methodology of the proposed soft knee rehabilitation system. We first provide a system-level overview of the rehabilitation scenario and the closed-loop information flow. We then describe the exoskeleton hardware design, including the origami-inspired pneumatic actuator, wearable integration, sensing, signal preprocessing, and practical safety measures. Finally, we present the human--robot coupled Koopman model and the KMPC formulation used for real-time assistive control.
\subsection{System Overview}
We consider a seated knee rehabilitation scenario in which the participant wears the soft knee exoskeleton while seated. The knee joint angle is measured in real time and mapped to a cursor displayed on a graphical user interface. The participant is instructed to track a prescribed reference trajectory under two rehabilitation modes. In passive mode, the participant remains relaxed and the knee motion is mainly driven by the soft exoskeleton. In active mode, the participant voluntarily contributes to knee motion while receiving pneumatic assistance from the exoskeleton.

\begin{figure*}[!htpb]
        \centering
        \includegraphics[width=0.8\textwidth]{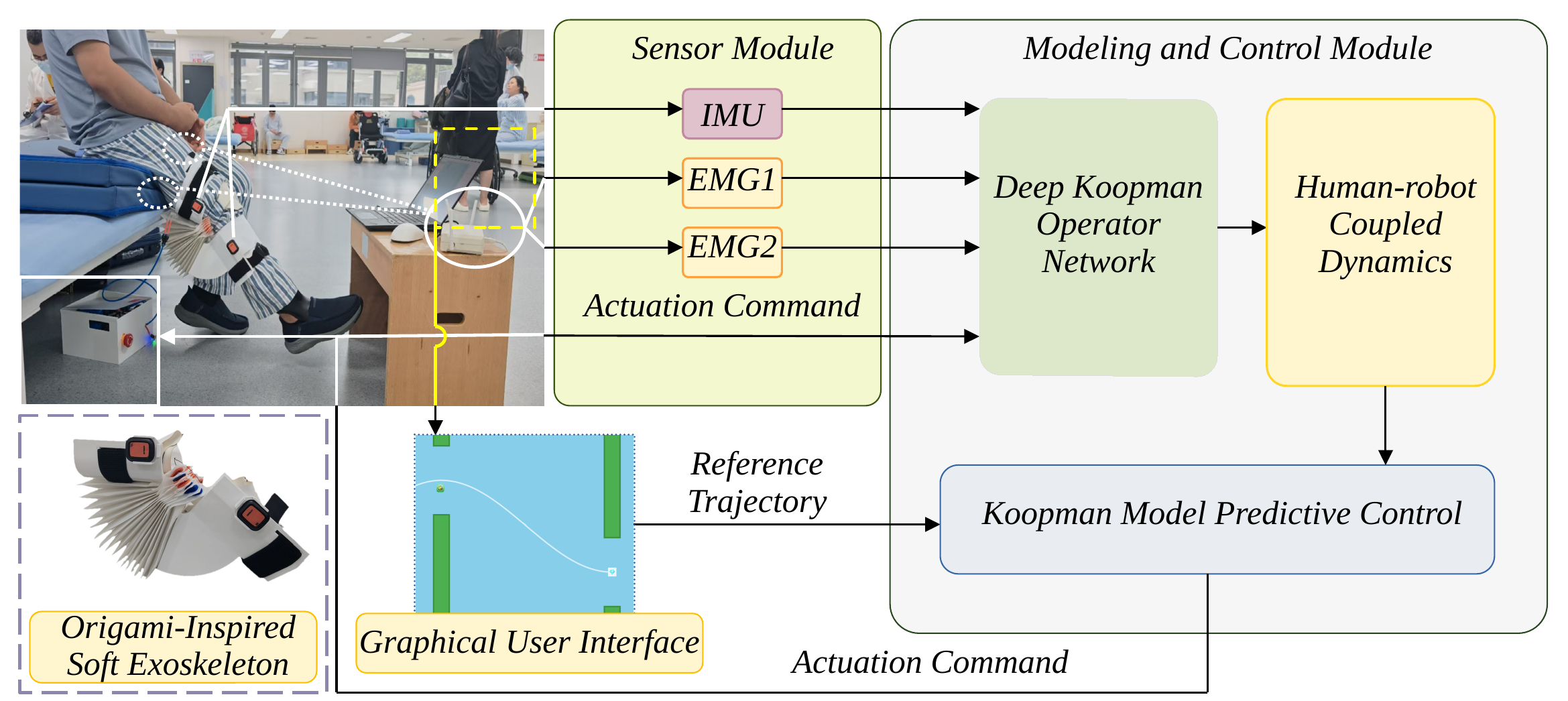}
        \caption{System overview of the proposed EMG-informed Koopman-MPC framework for soft knee rehabilitation. The framework integrates pneumatic assistance, IMU/EMG sensing, subject-specific Koopman modeling, and KMPC-based online control.}
        \label{fig:system_overview}
\end{figure*}

As illustrated in Fig.~\ref{fig:system_overview}, the proposed framework forms a closed-loop rehabilitation system that integrates pneumatic assistance, multimodal sensing, subject-specific modeling, and predictive control. The origami-inspired pneumatic actuator, together with the wearable fixation structures, provides assistive torque to the knee joint, while the IMU and EMG sensors measure the knee joint angle and neuromuscular activation during rehabilitation training. After signal preprocessing and delay alignment, the measured angle, EMG signals, and control inputs are used to learn a subject-specific Deep Koopman model of the human--robot coupled dynamics. During online control, the KMPC controller uses the learned model, the reference trajectory, and the EMG input to compute the actuation command for the pneumatic drive system in real time.
\subsection{Exoskeleton Design and Sensing}
\subsubsection{The Pneumatic Actuator}
\begin{figure}[!htpb]
        \vspace{-.3cm}
        \centering
        \includegraphics[width=0.5\textwidth]{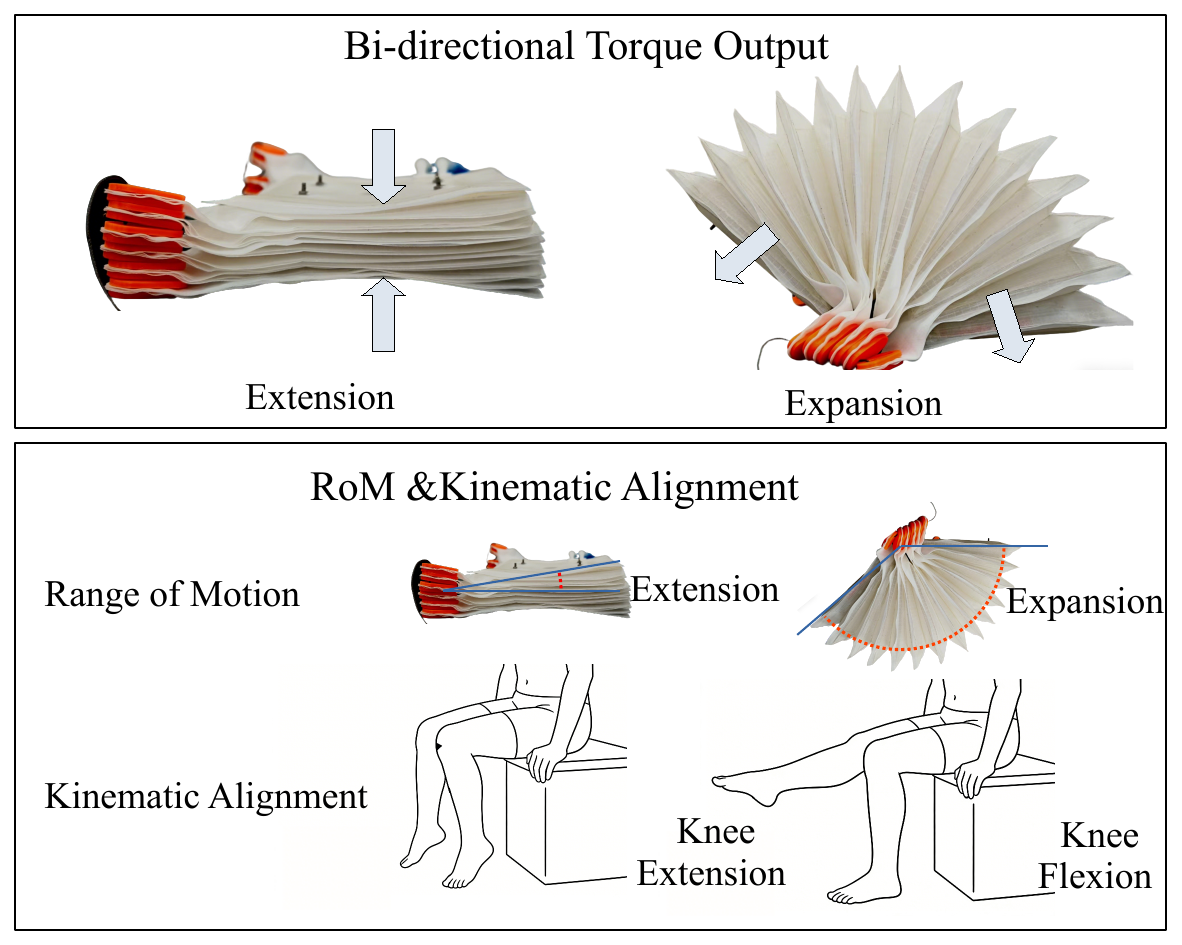}
        \caption{The origami-inspired pneumatic actuator illustrating bi-directional torque output, large Range of Motion (RoM), and kinematic alignment with the knee joint.}
        \label{fig:origami_actuator}
\end{figure}

The system employs an origami-inspired pneumatic actuator developed in \cite{10361525}.
Following the design and optimization methodology in \cite{10361525}, we adapt the origami-inspired pneumatic actuator to the mechanical and ergonomic requirements of seated knee rehabilitation, resulting in a portable and easy-to-wear soft knee rehabilitation exoskeleton.
The re-designed actuator consists of a quadrangular-expanded origami frame fabricated from Polyvinyl Chloride (PVC) sheets and embedded within a flexible air bladder. The origami skeleton geometrically constrains the deformation of the air bladder, which significantly improves stability under negative pressure and transforms the actuator behavior from conventional nonlinear bulging to a near-linear folding and unfolding motion. This soft--rigid integration enables a high contraction ratio and improved actuation efficiency. As shown in Fig.~\ref{fig:origami_actuator}, the actuator is capable of generating bi-directional torque and provides an optimized range of motion aligned with knee joint kinematics, making it well suited for natural flexion and extension during rehabilitation training.

\begin{figure}[!htpb]
        \vspace{-.3cm}
        \centering
        \includegraphics[width=0.48\textwidth]{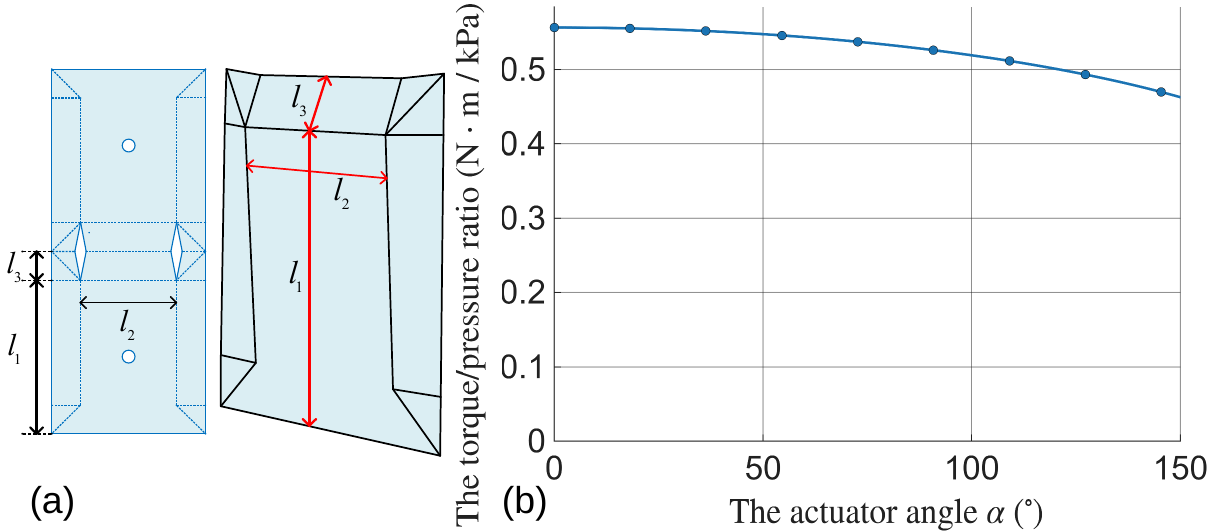}
        \caption{(a) Parameterization of the origami-inspired actuator \cite{10361525}. (b) Torque-to-pressure ratio as a function of actuator angle $\alpha$ with 16 air chambers, computed using $l_1 = 101$ mm, $l_2 = 70$ mm, and $l_3 = 20$ mm.}
        \label{fig:torque_coefficient}
\end{figure}

In this work, the actuator design is further adapted and optimized for knee rehabilitation based on the design and optimization methodology in \cite{10361525}. More specifically, the geometric parameters are redesigned to better match the anatomical structure and range of motion of the human knee. Consequently, the number of air chambers is set to 16 so that the actuator can maintain a favorable torque-to-pressure ratio over the knee-rehabilitation range of motion. As shown in Fig.~\ref{fig:torque_coefficient}(b), the 16-chamber configuration maintains a relatively high torque-to-pressure ratio over an angular range from $0^\circ$ to $100^\circ$, which is important for supporting knee flexion and extension during seated rehabilitation.
Moreover, the resulting lightweight design weighs approximately $0.7\,\mathrm{kg}$ and delivers an output torque of $17\,\mathrm{N\cdot m}$ at an operating condition of $30\,\mathrm{kPa}$ and $20^\circ$ joint angle.

The pneumatic actuator is then integrated into the wearable exoskeleton system.
Direct contact between the soft pneumatic actuator and the human leg can lead to uneven force transmission, resulting in adverse reaction forces and undesirable actuator deformations. To mitigate these issues, two 3D-printed connectors, shown in Fig.~\ref{fig:exoskeleton_integration}(d), are attached to both ends of the actuator. These connectors provide sufficient structural stiffness and facilitate a more uniform distribution of assistive forces across the thigh and lower leg.
In addition, the predefined orientation of the 3D-printed connectors shifts the actuator's high torque-to-pressure operating range to better align with the knee range of motion required for rehabilitation training.

\subsubsection{Wearable integration and fixation}
\begin{figure}[!htpb]
        \vspace{-.3cm}
        \centering
        \includegraphics[width=0.5\textwidth]{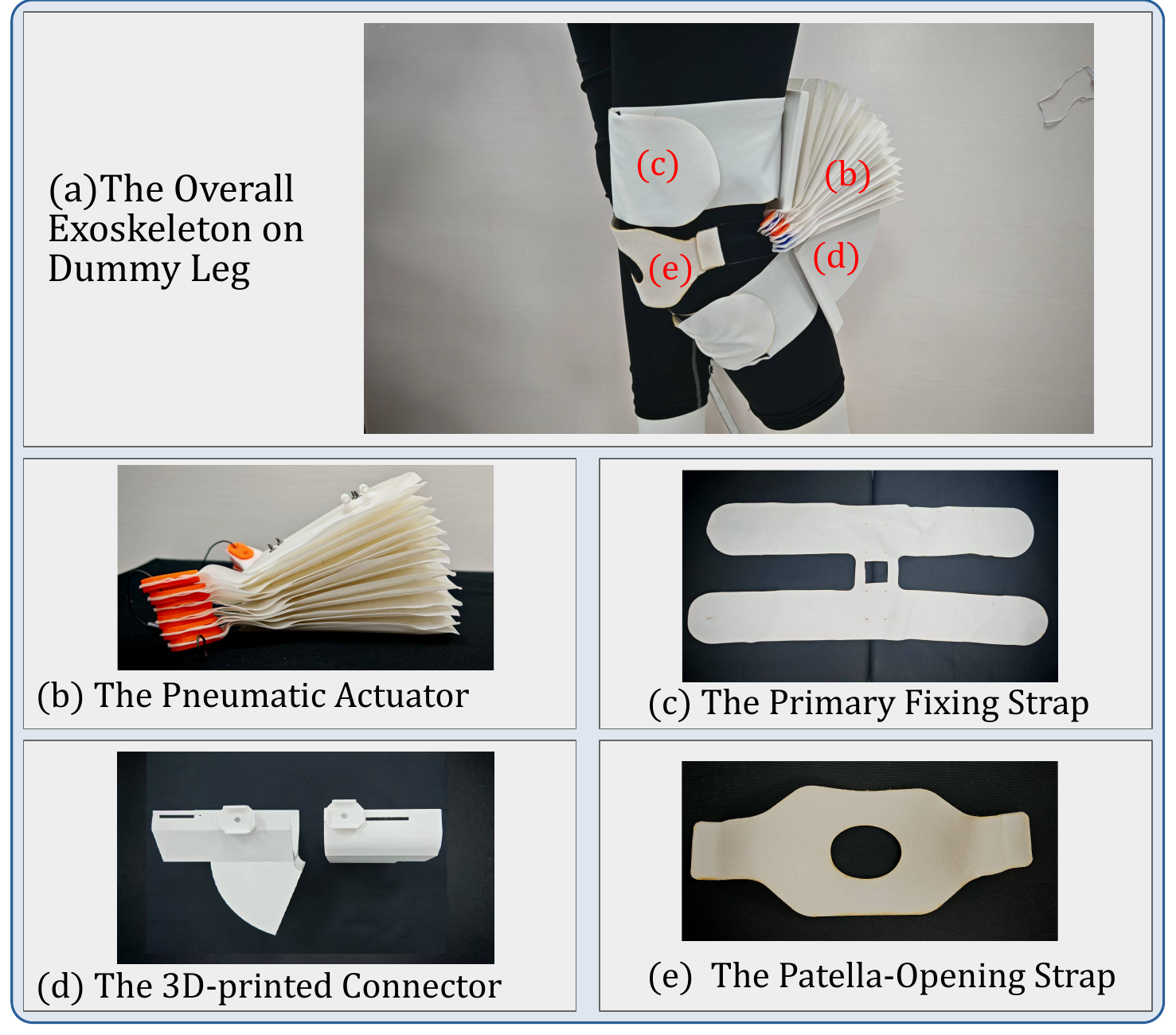}
        \caption{The overall exoskeleton design and its main components. (a) Assembled exoskeleton mounted on a dummy leg. (b) Pneumatic actuator. (c) Primary fixing strap. (d) 3D-printed connector. (e) Patella-opening strap.}
        \label{fig:exoskeleton_integration}
\end{figure}

To further securely attach the exoskeleton to the user, a primary fixing strap, shown in Fig.~\ref{fig:exoskeleton_integration}(c), is employed to fasten the connectors to the thigh and lower leg. This fastening mechanism allows rapid adjustment to accommodate different leg sizes while preventing relative slippage between the actuator and the limb. Furthermore, a patella-opening strap, shown in Fig.~\ref{fig:exoskeleton_integration}(e), is connected to the kneecap and linked to the actuator's string holders on both sides. These string holders are reinforced by 3D-printed components (the orange components shown in Fig.~\ref{fig:exoskeleton_integration}(b)), which prevent actuator center drift relative to the knee joint and improve force transmission efficiency. These connector and strap structures are designed to distribute the assistive torque over a larger contact area on the thigh and shank, reduce localized stress concentration, improve fixation stability, and help reduce relative slippage between the exoskeleton and the limb.

\subsubsection{The Sensing Module}
\begin{figure}[!htpb]
        \vspace{-.3cm}
        \centering
        \includegraphics[width=0.45\textwidth]{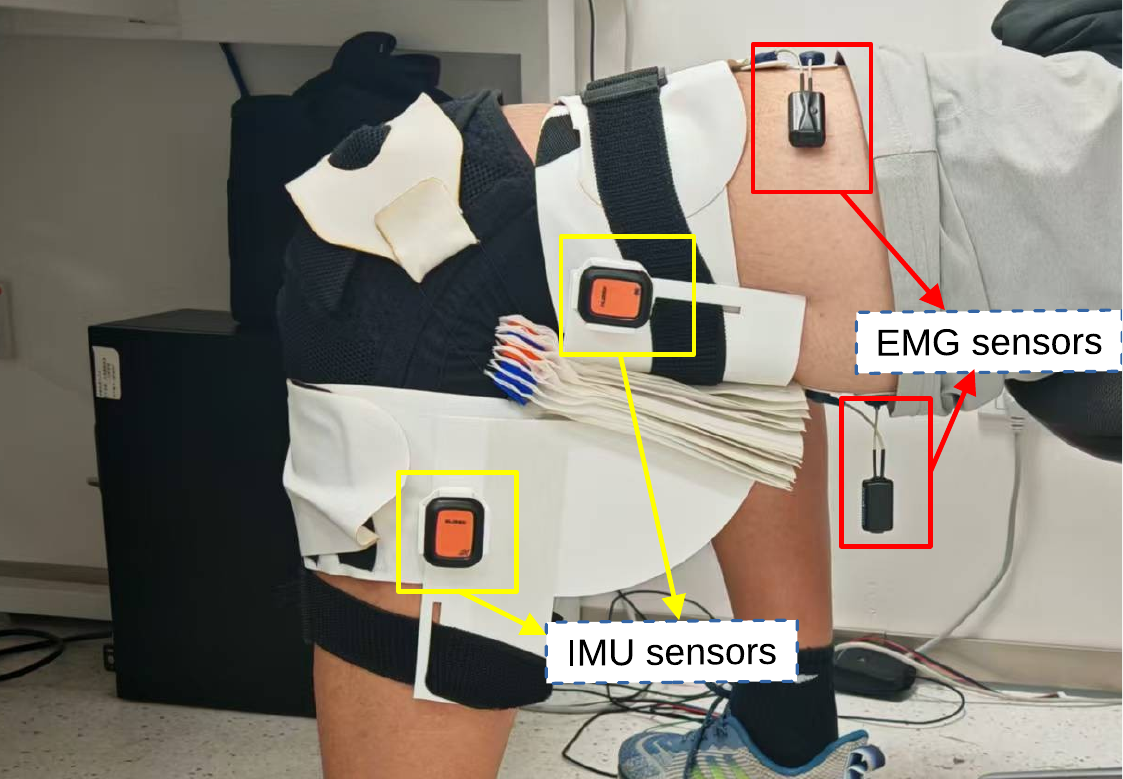}
        \caption{Placement of EMG and IMU sensors on the lower limb. EMG sensors are placed on the thigh muscles, and IMUs are mounted on the thigh and lower leg to measure joint angles.}
        \label{fig:sensor_placement}
\end{figure}

As illustrated in Fig.~\ref{fig:sensor_placement}, two Inertial Measurement Units (IMUs) are mounted on the 3D-printed connectors to measure the knee joint angle $x_k$. The measured joint angle is related to the actuator angle $\alpha_k$ by $x_k = \alpha_k + \Delta$,
where $\Delta$ denotes the fixed offset introduced by the connector geometry.

Moreover, to capture the muscle activation and voluntary neuromuscular contribution during knee motion, electromyography (EMG) sensors are placed on the rectus femoris (EMG1) and biceps femoris (EMG2), which are the primary muscles responsible for knee extension and flexion, respectively. The raw EMG signals are sampled at 2000~Hz and are subject to significant noise contamination. To ensure reliable Koopman model identification, each EMG signal undergoes the following preprocessing steps \cite{6198287}:
\begin{itemize}
    \item Band-pass filtering using a 4\textsuperscript{th}-order Butterworth filter with cutoff frequencies of 20~Hz and 450~Hz;
    \item Computation of the Root-Mean-Square (RMS) value using a sliding window of 20 samples;
    \item Low-pass filtering of the RMS signal using a 1\textsuperscript{st}-order Infinite Impulse Response (IIR) filter with a 3~Hz cutoff frequency to extract the signal envelope;
    \item Normalization based on the maximum RMS value recorded offline, with dynamic updates when the current RMS exceeds the stored maximum, yielding normalized signals in the range $[0,1]$.
\end{itemize}

After preprocessing, the EMG signals are synchronized with the IMU measurements, and both are updated at the 50~Hz frequency.

\subsubsection{The Pneumatic Control System Hardware Architecture}
Achieving responsive and stable actuation for knee rehabilitation requires the pneumatic system to deliver sufficient airflow under varying load conditions. To meet this requirement, the system employs two Kamoer KZP20-B24 diaphragm pumps, each with a maximum power of 50~W, to drive the pneumatic actuator. A directional control valve is integrated to regulate the airflow direction. By switching the valve state between inflation and deflation, the system enables bi-directional assistance for knee flexion and extension. In addition, we use a 6000mAh battery to power the system, which can support over one hour of continuous operation under typical experimental conditions. 

\begin{figure}[!htpb]
        \vspace{-.3cm}
        \centering
        \includegraphics[width=0.5\textwidth]{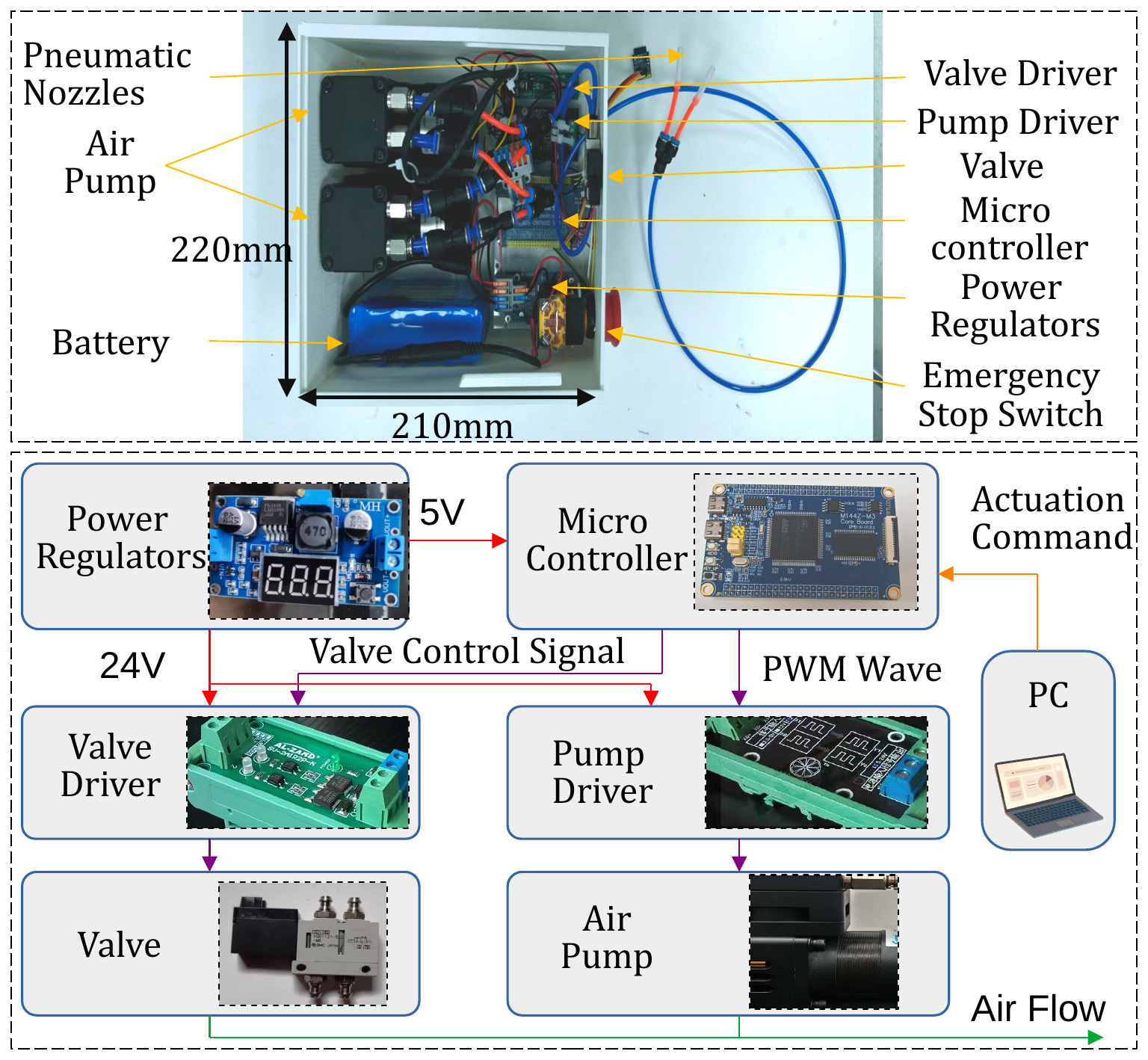}
        \caption{Hardware architecture of the pneumatic drive and control unit. The diagram shows the integration of the host computer, microcontroller, power module, pump and pump driver, valve and valve driver, together with the corresponding control signals and airflow paths.}
        \label{fig:system_architecture}
\end{figure}

As illustrated in Fig.~\ref{fig:system_architecture}, the pneumatic control system consists of a power supply module, a microcontroller, a pump driver module, and a valve driver module. The power supply provides regulated 24~V and 5~V outputs to the system components. The microcontroller serves as the low-level execution layer, receiving high-level actuation commands from the host computer and generating the corresponding Pulse-Width Modulation (PWM) wave for hardware actuation.
In addition, the airflow direction is controlled through discrete switching of the directional valve, while the airflow rate is controlled via adjusting the pump motor speed by PWM wave regulation.
This hardware architecture allows independent control of the actuation direction and magnitude, thereby providing the necessary flexibility for implementing the proposed control framework.
The entire integrated box shown in Fig.~\ref{fig:system_architecture} is 220 mm$\times$210 mm$\times$130 mm in size, and weighs 1.3 kg.

Moreover, several practical safety measures are incorporated in the pneumatic drive system and controller implementation. A physical emergency stop button immediately cuts off power to the entire device when activated, and the system ceases inflation in the event of a power failure. The pneumatic nozzles are designed to be easily detachable, so that the air pipes can be quickly disconnected from the actuator in an emergency and the actuator pressure can be released to atmospheric level. In addition, a software-level hard joint-angle limit is implemented in the controller; once the measured knee angle exceeds the preset safety threshold of $170^\circ$, the controller switches to a protective deflation command to prevent over-inflation. These mechanisms help reduce the risk of uncontrolled inflation during rehabilitation experiments. The system is also capable of continuous operation for over one hour under typical experimental conditions.

\subsection{Human--Robot Coupled Koopman Modeling}
\label{sec:koopman_model_and_training}
\begin{figure*}[!htpb]
    \centering
    \includegraphics[width=1\textwidth]{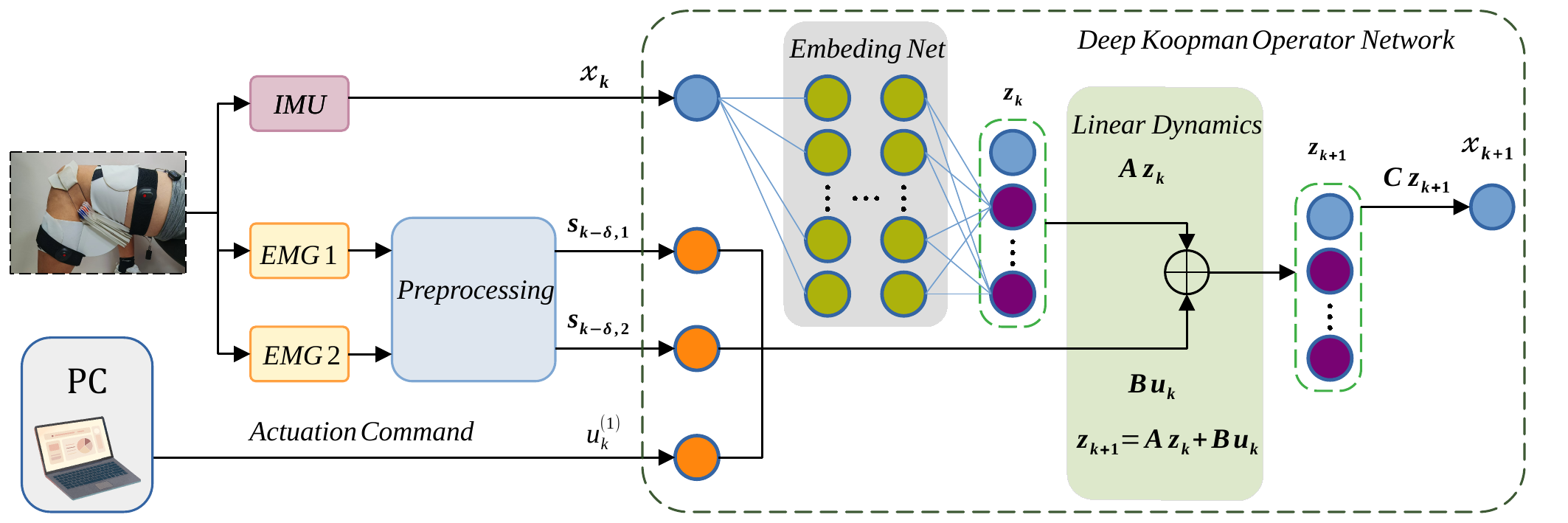}
    \caption{Deep Koopman operator network framework. The encoder maps the physical state $x_k$ into learned features, which are concatenated with the original state to form the lifted state $\mathbf{z}_k$. The lifted state evolves linearly under the influence of the delayed EMG input $\mathbf{s}_{k-\delta}$ and the control input $u^{(1)}_k$, and the physical state is recovered through a linear output map.}
    \label{fig:Deep_Koopman_Network}
\end{figure*}

Origami-inspired pneumatic actuators exhibit strongly nonlinear dynamics due to the intrinsic compliance of soft materials and complex geometric deformation patterns. When integrated into a wearable exoskeleton and coupled with the human musculoskeletal system, the resulting human--robot dynamics become highly complex. While simplified pressure-to-angle mappings or rigid-body-based models can capture certain aspects of the actuator behavior, they are generally insufficient for precise control. Moreover, obtaining accurate musculoskeletal models in clinical settings is impractical, as the key parameters such as limb mass distribution, inertia, muscle strength, reaction latency, and joint stiffness vary significantly across individuals.

To address these challenges, a data-driven modeling approach based on Koopman operator theory is adopted. The system is formulated in discrete time with a sampling period of $\Delta t = 20\,\mathrm{ms}$. 
The knee joint angle $x_k$ at time step $k$ is defined as the system state. The input vector is given by
\[
\mathbf{u}_k = \big[u^{(1)}_k,\; \mathbf{u}^{(2)\top}_k \big]^\top \in \mathbb{R}^3,
\]
where $u^{(1)}_k$ denotes the actuation command that controls the pump and valve, which determines the assistive torque generated by the exoskeleton, and is treated as the control input in the proposed framework. $\mathbf{u}^{(2)}_k \in \mathbb{R}^{2}$ represents the EMG signals collected from the two muscle channels. Notably, the EMG signals reflect neuromuscular activation and implicitly determine the muscle-generated torque applied to the knee joint.

Since EMG signals precede actual muscle contraction, the muscle contribution that affects the state evolution is modeled using a time-shifted EMG signal. More specifically, let $\mathbf{s}_k \in \mathbb{R}^{m-1}$ denote the EMG measurement at time step $k$, and we let $\mathbf{u}^{(2)}_k := \mathbf{s}_{k-\delta}$, where $\delta$ represents the neuromuscular delay. Both $u^{(1)}_k$ and $\mathbf{u}^{(2)}_k$ are therefore treated as inputs to the human--robot coupled dynamics.

Next, a lifting function $\phi : \mathbb{R} \rightarrow \mathbb{R}^d$ is introduced to map the knee angle into a high-dimensional lifted state $\mathbf{z}_k = \phi(x_k)$. In particular, the lifting function is implemented using a Multi-Layer Perceptron (MLP) encoder, as illustrated in Fig.~\ref{fig:Deep_Koopman_Network}. The encoder consists of three hidden layers with 128 neurons each and  ReLU activation functions, producing a 96-dimensional feature vector. The lifted state is constructed by concatenating the original state and the encoded features. Hence the lifted state dimension $d$ is 97.
Then the lifted dynamics are assumed to evolve linearly according to
\begin{equation}
    \mathbf{z}_{k+1} = \mathbf{A}\mathbf{z}_k + \mathbf{B}\mathbf{u}_k,
    \label{eq:koopman_linear_dynamics}
\end{equation}
where $\mathbf{A} \in \mathbb{R}^{d \times d}$ and $\mathbf{B} \in \mathbb{R}^{d \times 3}$ form a finite-dimensional approximation of the Koopman operator. 

The knee angle is recovered via the linear output map $x_k = \mathbf{C}\mathbf{z}_k$, with $\mathbf{C} = [1\ \mathbf{0}] \in \mathbb{R}^{1 \times d}$. This formulation yields an output-tracking control problem for the lifted linear system, where $u^{(1)}_k$ serves as the control input and $\mathbf{u}^{(2)}_k$ acts as a known feedforward signal. 

To train the lifting function $\phi$ and the Koopman matrices $(\mathbf{A}, \mathbf{B})$, data are collected by asking the participant to track a reference trajectory while wearing the soft exoskeleton. In particular, the soft exoskeleton is controlled by a baseline PID controller. The resulting dataset is constructed as
\begin{align}
\mathcal{D} := \big\{(x_k^i, \mathbf{u}_k^i, x_{k+1}^i)\,|\, k = 0,\ldots,N_b-1,\ i = 1,\ldots,M \big\},
\label{eq:dataset}
\end{align}
where the data are organized into $M$ trajectory batches of length $N_b = 16$. To improve numerical conditioning during the training process, the joint angle measurements are scaled by a factor of 20, and the actuation commands are normalized to the range $[-1,1]$.

A multi-step prediction loss
\begin{align*}
\mathcal{L}_{\text{pred}} :=
\sum_{i=1}^{M}\sum_{k=1}^{N_b}
\gamma^{k-1}
\big\|
\hat{\mathbf{z}}_k
-
\phi(x_k^i)
\big\|^2
\end{align*}
is employed to train the model,
where $\gamma = 0.9$ is a temporal discount factor. 

The network parameters are optimized using the Adam optimizer with a Noam learning rate schedule \cite{Vaswani2017}. The model is trained for 8000 iterations with a batch size of 10.

\subsection{KMPC and Real-Time Implementation}\label{sec: real_time_implementation}
Based on the identified Koopman operator model of the human--robot coupled system, a Koopman Model Predictive Control (KMPC) framework is formulated to regulate the assistive actuation of the soft exoskeleton. To facilitate control design, the input matrix in \eqref{eq:koopman_linear_dynamics} is decomposed according to the robot and human contributions, which yields
\begin{equation}
\mathbf{z}_{k+1}
=
\mathbf{A}\mathbf{z}_k
+
\mathbf{B}_1 u^{(1)}_k
+
\mathbf{B}_2 \mathbf{u}^{(2)}_k,
\label{eq:split_inputs}
\end{equation}
where $\mathbf{B} = [\mathbf{B}_1\ \mathbf{B}_2]$, with $\mathbf{B}_1 \in \mathbb{R}^{d \times 1}$ corresponding to the control input generated by the soft exoskeleton and $\mathbf{B}_2 \in \mathbb{R}^{d \times (m-1)}$ associated with the human muscle activation signals.

The control objective is to drive the human--robot coupled system \eqref{eq:split_inputs} to track a prescribed reference knee angle trajectory $\{x_k^r\}_{k=1}^{\infty}$. At each time step $k$, the KMPC computes the optimal control sequence by solving the following finite-horizon optimization problem:
\begin{equation}
\begin{aligned}
\min_{\{\mathbf{z}_{k+t|k}\},\{u^{(1)}_{k+t|k}\}} 
& \sum_{t=1}^{N}
q \, \| \mathbf{C}\mathbf{z}_{k+t|k} - x^r_{k+t} \|^2
+
r \, \| u^{(1)}_{k+t-1|k} \|^2
 \\
\text{s.t.}\quad
& \mathbf{z}_{k+t+1|k}
=
\mathbf{A}\mathbf{z}_{k+t|k}
+
\mathbf{B}_1 u^{(1)}_{k+t|k}
+
\mathbf{B}_2 \mathbf{u}^{(2)}_{k+t}, \\
& u_{\min} \le u^{(1)}_{k+t|k} \le u_{\max}, \quad t = 0, \dots, N-1,
\end{aligned}
\label{eq:kmpc_optimization}
\end{equation}
where $q = 1$ and $r = 0.25$ weight the tracking error and control effort, respectively. In addition, we set the bounds as $u_{\min} = -1$ and $u_{\max} = 1$ to align with the physical definition of the PWM wave duty cycle.

Notably, in \eqref{eq:kmpc_optimization}, the human contribution $\mathbf{u}^{(2)}_{k+t}$, derived from time-shifted EMG signals, is treated as a known feedforward input, while the robot assistance $u^{(1)}_k$ is optimized to compensate for both the desired task and the user's voluntary motion. We refer to this mechanism as EMG-informed assistance: the EMG signals inform the predicted human contribution, and the KMPC accounts for this contribution when optimizing robot actuation to balance tracking error and actuation effort. Due to the linear lifted dynamics and quadratic cost function, the KMPC problem reduces to a convex quadratic program. 
Namely, let $\mathbf{\zeta}_k:=[\mathbf{z}_{k|k}^\top,\mathbf{z}_{k+1|k}^\top,\ldots,\mathbf{z}_{k+N|k}^\top,u^{(1)}_{k|k},\ldots,u^{(1)}_{k+N-1|k}]^\top$. Then the optimization problem \eqref{eq:kmpc_optimization} can be written into the following compact form
\begin{align*}
        \min_{\mathbf{\zeta}_k}  &\; \mathbf{\zeta}_k^\top \mathbf{H} \mathbf{\zeta}_k + \mathbf{c}^\top \bm{\zeta}_k \\
        &\;\mathbf{G}\mathbf{\zeta}_k = \mathbf{\eta}^{(2)}\\
        &\;\mathbf{E}\mathbf{\zeta}_k \le \mathbf{\xi},
\end{align*}
where $\mathbf{H}=\diag(q\mathbf{C}^\top C,\allowbreak \ldots,\allowbreak q\mathbf{C}^\top C,\allowbreak r,\allowbreak \ldots,\allowbreak r)$, $\mathbf{c}^\top=[-2q\mathbf{x}_{k+1}^{r,\top}\mathbf{C},\allowbreak \ldots,\allowbreak -2q\mathbf{x}_{k+N}^{r,\top}\mathbf{C},\allowbreak 0,\allowbreak \ldots,\allowbreak 0]^\top$, $\mathbf{\xi}=[u_{\max},\allowbreak \ldots,\allowbreak u_{\max},\allowbreak -u_{\min},\allowbreak \ldots,\allowbreak -u_{\min}]^\top$, $\mathbf{E}=[\mathbf{0}\,|\,\diag(\mathbf{I},\allowbreak \ldots,\allowbreak \mathbf{I},\allowbreak \mathbf{-I},\allowbreak \ldots,\allowbreak \mathbf{-I})]$, and
\begin{align*}
        \mathbf{G} = \begin{bmatrix}
                \mathbf{A} & -\mathbf{I} & & &\mathbf{B}_1\\&\ddots &\ddots & & &\ddots \\&&\mathbf{A} &-\mathbf{I} & & & \mathbf{B}_1
        \end{bmatrix}.
\end{align*}
By precomputing the matrices and vectors in the compact formulation, the resulting quadratic program can be solved efficiently using a standard numerical solver. In this work, we use the Operator Splitting Quadratic Program (OSQP) solver \cite{stellato2018osqp}.

Such formulation enables subject-specific assistive control strategies, as individual biomechanical properties and interaction dynamics are implicitly captured by the learned Koopman model. By explicitly accounting for human muscle activation rather than treating it as an external disturbance, the proposed KMPC framework supports EMG-informed assistance for knee rehabilitation across different users.

\section{Experimental Protocol}
In the experiments, all controllers are run on a ThinkPad T14p Gen 2 equipped with an Intel Core Ultra 9 185H CPU (up to 5.1 GHz) and 32 GB of RAM.
The experiments are conducted on both healthy participants and stroke patients. All experimental procedures are approved by the Medical Ethics Committee of the Seventh Affiliated Hospital of Sun Yat-sen University (Approval No.~SHSYSU202501103240110). All participants provided written informed consent prior to participation and were informed of their right to withdraw at any time.

\subsubsection{The Participants}

\begin{itemize}
    \item \textbf{Healthy Participants (P1--P5):} five male participants aged between 23 and 26 years.
    \item \textbf{Patients (S1--S11):} 11 stroke patients (7 males and 4 females) aged between 31 and 54, with residual lower-limb motor function sufficient to perform limited knee flexion and extension, and with normal cognitive abilities. All recruited patients are within Brunnstr\"om stages 3--5. No EMG-based screening, predefined signal-quality threshold, or clinical EMG assessment is used for patient inclusion.
\end{itemize}

\begin{figure*}[!htpb]
    \centering
    \includegraphics[width=\textwidth]{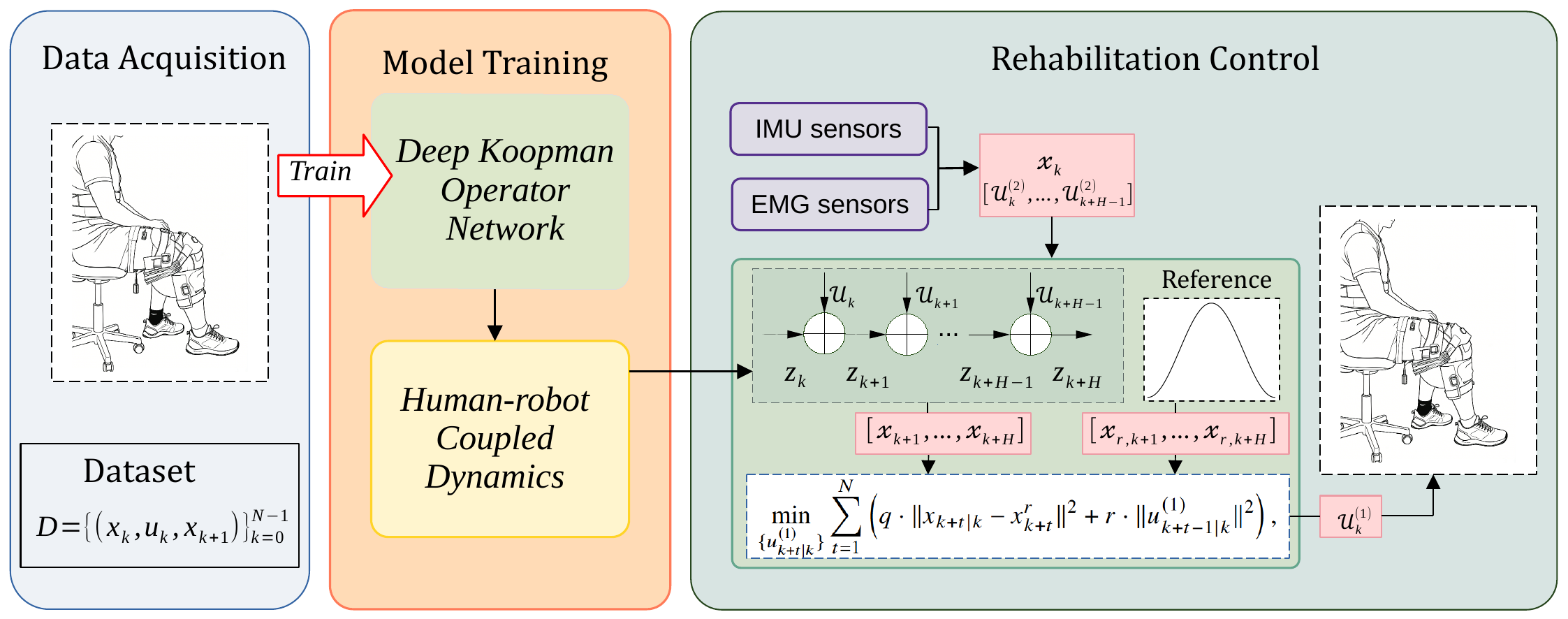}
    \caption{Overview of the experimental framework, including data acquisition, model training, and rehabilitation control.}
    \label{fig:Whole_control_framework}
\end{figure*}

\begin{table}[!htpb]
\centering
\caption{Participant Demographics}
\label{tab:attributes_participants}
\footnotesize
\renewcommand{\arraystretch}{1.2}
\setlength{\tabcolsep}{3pt}
\begin{tabular*}{\linewidth}{@{} @{\extracolsep{\fill}} c c c c c c @{}}
\toprule
\textbf{ID} & \textbf{Gender} & \makecell{\textbf{Age}\\\textbf{(y)}} & \makecell{\textbf{Weight}\\\textbf{(kg)}} & \makecell{\textbf{Height}\\\textbf{(cm)}} & \makecell{\textbf{Brunnstr\"om}\\\textbf{Stage}} \\
\midrule
 P1 & male & 24 & 88 & 188 & / \\
 P2 & male & 25 & 72 & 185 & / \\
 P3 & male & 26 & 85 & 181 & / \\
 P4 & male & 23 & 83 & 183 & / \\
 P5 & male & 25 & 66 & 172 & / \\
 S1 & female & 31 & 70 & 163 & 4 \\
 S2 & male & 48 & 54 & 168 & 4 \\
 S3 & female & 54 & 45 & 160 & 5 \\
 S4 & male & 33 & 83 & 172 & 4 \\
 S5 & male & 39 & 66 & 166 & 4 \\
 S6 & male & 50 & 74 & 168 & 3 \\
 S7 & female & 38 & 55 & 161 & 4 \\
 S8 & male & 46 & 80 & 173 & 4 \\
 S9 & male & 36 & 82 & 172 & 4 \\
 S10 & female & 41 & 78 & 165 & 5 \\
 S11 & male & 52 & 71 & 170 & 3  \\
\bottomrule
\end{tabular*}
\end{table}

The participant demographics are summarized in Tab.~\ref{tab:attributes_participants}.

\subsubsection{Experimental Procedure}
\label{sec:experimental_procedure}

The experimental procedure consists of three stages: data acquisition, Koopman model training, and rehabilitation training, as illustrated in Fig.~\ref{fig:Whole_control_framework}.

\begin{figure}[!htpb]
\centering
\includegraphics[width=1\columnwidth]{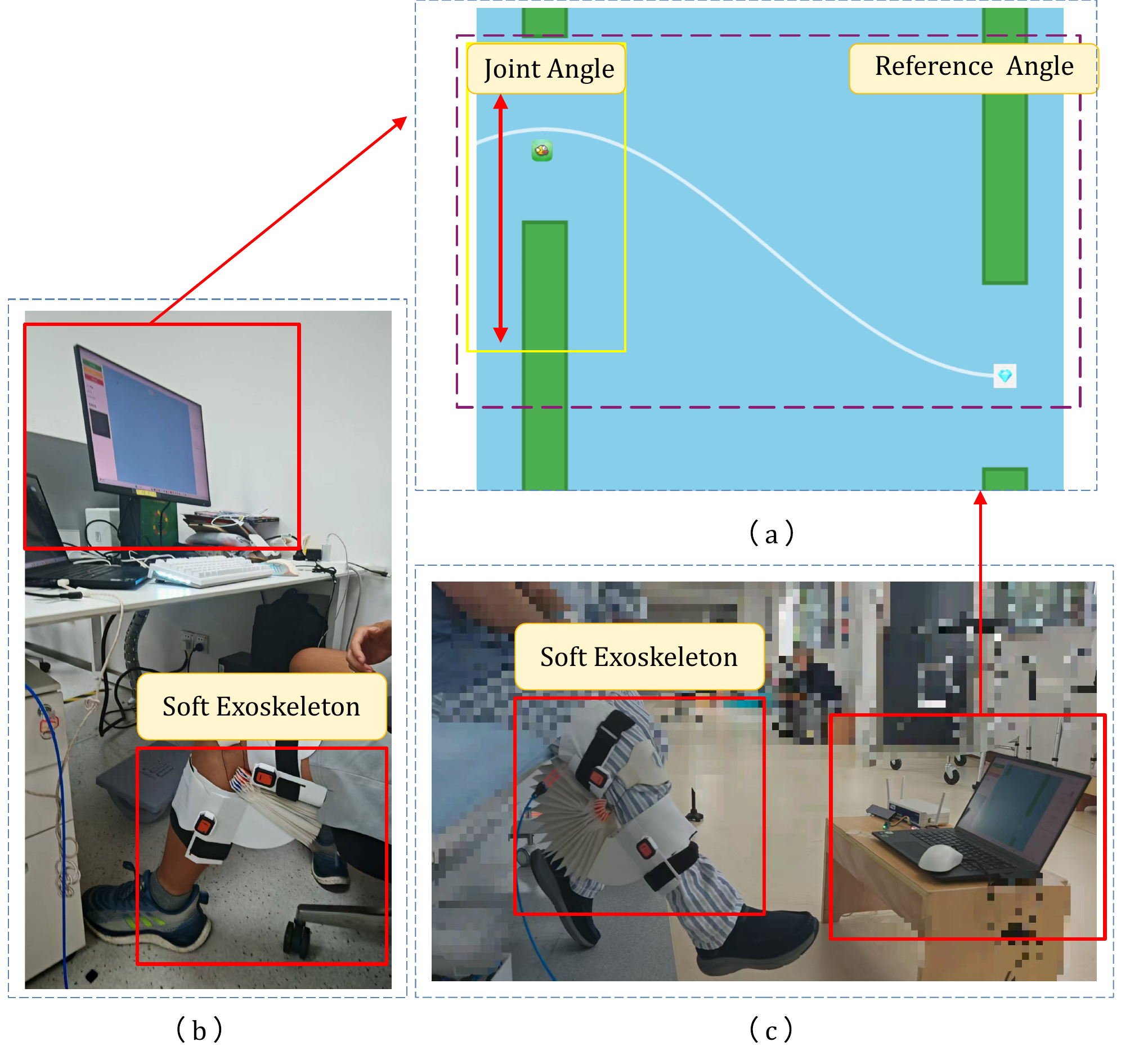}
\caption{(a) GUI for knee angle tracking. (b) Experimental setup for healthy participants. (c) Experimental setup for patients.}
\label{fig:Experiment photo}
\end{figure}

During the experiments, the knee joint angle is mapped to the vertical position of a cursor in a Graphical User Interface (GUI), as shown in Fig.~\ref{fig:Experiment photo}(a). Participants are instructed to track a reference trajectory displayed in the GUI by moving their knee under assistance from the soft exoskeleton. Experimental setups for the healthy participants and the patients are shown in Fig.~\ref{fig:Experiment photo}(b) and (c), respectively.

Due to the differences in motor capability and clinical rehabilitation requirements, different reference trajectories are employed for healthy participants and patients:
\begin{itemize}
    \item \textbf{Healthy Participants:} a sinusoidal reference with a frequency of 0.032~Hz and an amplitude that corresponds to knee angles between $90^{\circ}$ and $120^{\circ}$.
    \item \textbf{Patients:} a sinusoidal reference with a frequency of 0.04~Hz and an amplitude that corresponds to knee angles between $90^{\circ}$ and $160^{\circ}$. The larger motion range is selected to accommodate the clinical rehabilitation needs, while the trajectory remains slow and smooth to ensure the patients' safety.
\end{itemize}

During the data acquisition stage, the soft exoskeleton is controlled by a conventional PID controller while the participants track the reference trajectories. Data are collected under two rehabilitation modes:
\begin{itemize}
    
    \item \textbf{Passive Mode:} The participant remains relaxed, and knee motion is entirely driven by the exoskeleton.
    \item \textbf{Active Mode:} The participant actively contributes to the motion, and knee movement results from combined muscle activation and exoskeleton assistance.
\end{itemize}

Each training trial lasts 3 minutes, during which the knee joint angles, the EMG signals, and the control inputs are recorded to construct the dataset $\mathcal{D}$ described in Section~\ref{sec:koopman_model_and_training}. Data from both modes are pooled for model training. In addition, adequate rest periods are provided between trials, and the participants' comfort is continuously monitored.

Based on~\cite{doi:10.1152/japplphysiol.00135.2018}, the delay between neural activation and muscle force generation is approximately $100$--$200\,\mathrm{ms}$. In this study, a delay of $200\,\mathrm{ms}$ is adopted as a practical EMG-alignment choice. With the 20~ms sampling period, this delay corresponds to $\delta=10$ samples, and the KMPC prediction horizon is set to $N=10$.

Next, the personalized Koopman model is trained as described in Section~\ref{sec:koopman_model_and_training}, with a typical training time of approximately 20 minutes, which is compatible with the current clinical experimental protocol. Subsequently, rehabilitation training is conducted using the trained model and the KMPC. Each rehabilitation session lasts 3 minutes, and both passive and active rehabilitation performance are evaluated.

\subsubsection{Baseline Controllers and Evaluation Metrics}
To evaluate the performance of the proposed KMPC strategy, two baseline controllers are implemented for comparison: a conventional PID controller and ILC.
Since PID is the standard controller for soft exoskeletons and is also used for the initial data acquisition, it serves as the primary benchmark.
On the other hand, we choose ILC as a comparison benchmark because it is a well-established learning-based control strategy for repetitive rehabilitation tasks, and therefore provides a meaningful stronger benchmark for the rehabilitation setting considered in this work.
To ensure a fair comparison, all controllers are tested under the same reference trajectories, hardware platform, and 50 Hz control frequency. The PID controller uses fixed gains tuned for stable tracking across all subjects. The PID gains used for both data collection and baseline comparison are $K_p=3.0$, $K_i=0.1$, and $K_d=0.5$; these gains are selected in preliminary trials using classical sequential PID tuning and are kept fixed for all participants. The KMPC settings are also fixed across participants, including $N=10$, $q=1$, $r=0.25$, $u_{\min}=-1$, $u_{\max}=1$, and the OSQP solver; since the controlled output is the one-dimensional knee angle, the weights $q$ and $r$ are selected using sequential weight tuning to balance tracking performance and actuation effort. For ILC, we adopt a PD-type iterative learning update law by following \cite{7090671}. In particular, the feedforward term is first learned iteratively offline from PID-collected trajectory data for each subject and then applied online with PD feedback. The ILC comparison is conducted on stroke patients S8--S11, for whom additional ILC trials are available.

Finally, to quantitatively evaluate the accuracy of Deep Koopman model, we use the Mean Absolute Error (MAE) and the coefficient of determination ($R^2$) as the evaluation metrics.
In particular, for validation dataset \eqref{eq:dataset}, the $k$-step prediction MAE takes the form
\begin{align*}
    \mbox{MAE}(k) = \frac{1}{M}\sum_{i=1}^M |\hat{x}_k^i-x_k^i|,
\end{align*}
where $\hat{x}_k^i$ is the $k$-step prediction by Deep Koopman model based on the initial value $x_0^i$ and with the input sequence $\{\mathbf{u}_t^i\}_{t=0}^{k-1}$.
Moreover, we further use \emph{coefficient of determination} ($R^2$) to evaluate the overall predictive accuracy of the Koopman model across the entire prediction horizon. In particular, the $R^2$ takes the form  
\begin{align*}
R^2 &= 1 - \frac{\sum_{i=1}^M\sum_{k=1}^{N_b}\left(\hat{x}_k^i-x_k^i\right)^2}
{\sum_{i=1}^M\sum_{k=1}^{N_b}\left(x_k^i-\bar{x}\right)^2},
\end{align*}
where $\bar{x}:= \frac{1}{M\cdot N_b}\sum_{i=1}^M\sum_{k=1}^{N_b}x_k^i$ is the mean value of the measured knee joint angle across the entire validation dataset.

On the other hand, to quantitatively compare the controllers, we use the root mean square error (RMSE) to evaluate the tracking performance based on the entire dataset,
namely,
\begin{align*}
\text{RMSE} = \sqrt{\frac{1}{N}\sum_{k=1}^{N}\left(x_k^{r}-x_k\right)^2},
\end{align*}
where $N$ is the total number of time steps in the entire dataset.

Furthermore, we use a histogram to evaluate the distribution of computation time consumed by the KMPC controller at each time step. We also evaluate the effectiveness of the robotic assistance by comparing the RMS amplitudes of the EMG signals collected from Rectus Femoris and Biceps Femoris muscles under different training modes (no assistance, active and passive training modes).

\section{Results}
This section presents the experimental evaluation of the proposed human-robot coupled Koopman modeling and KMPC framework. We first assess the prediction accuracy of the learned deep Koopman model. We then evaluate the trajectory tracking performance of the personalized KMPC strategy against conventional PID and ILC baselines in both active and passive rehabilitation modes. Finally, we analyze the real-time computational efficiency and the effectiveness of the exoskeleton assistance in reducing the user's neuromuscular effort.

\subsection{Prediction Performance of the Deep Koopman Model}
Since the performance of the KMPC controller critically depends on the predictive accuracy of the learned dynamics, we first evaluate the prediction performance of the trained Koopman model for the human--robot coupled system. In addition, this evaluation also focuses on assessing the impact of incorporating the EMG signals as model inputs, which explicitly account for the human neuromuscular activity in the coupled dynamics.
For this purpose, the collected dataset is randomly split, with 80\% used for model training and the remaining 20\% reserved for testing. Then the prediction errors of two Koopman models are compared on the test set: one that incorporates the EMG signals as inputs and one that relies solely on kinematic information.

\subsubsection{Evaluation on the Healthy Participants}

\begin{figure}[!htpb]
\centering
    \begin{subfigure}{\columnwidth} 
    \centering
    \includegraphics[width=1\linewidth]{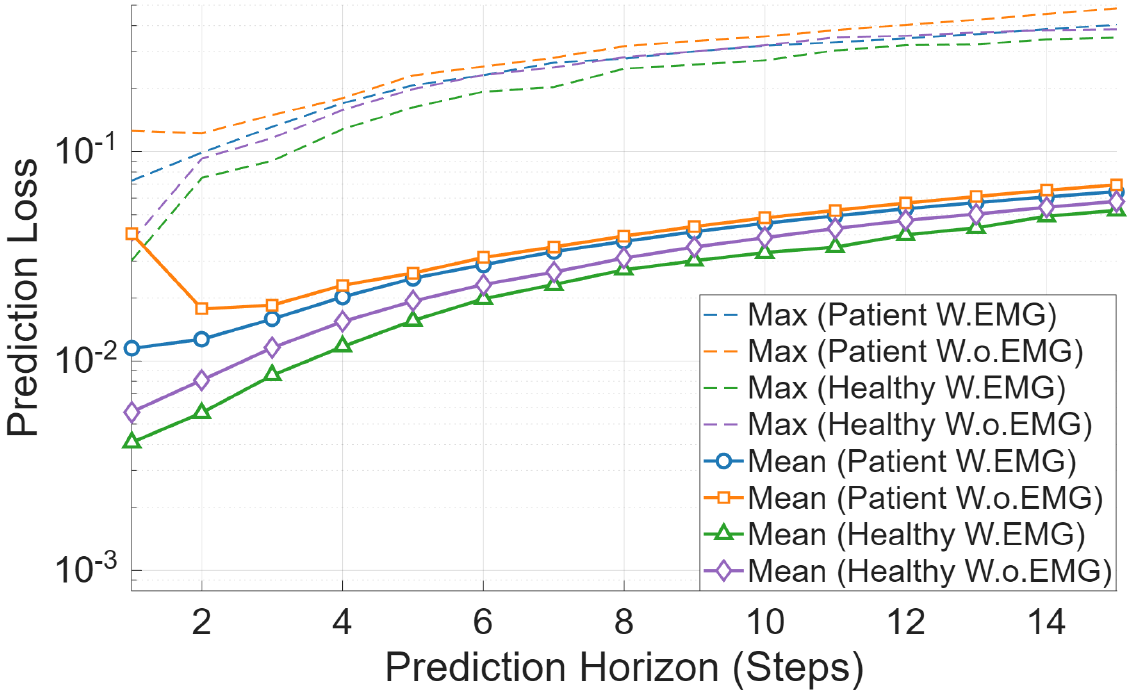}

    \caption{The mean prediction loss comparison between Koopman models with and without the EMG input for the healthy participants and patients.
    }
    \label{fig:emg_loss_comparison_healthy and patientss}
    \end{subfigure}
    \vspace{1em} 
    \begin{subfigure}{\columnwidth}
        \centering
        \includegraphics[width=1\linewidth]{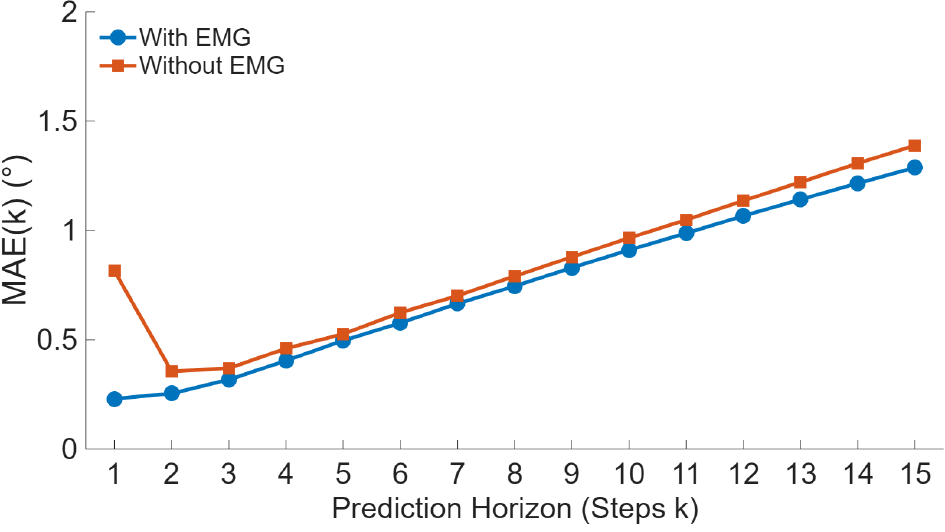}
    
        \caption{The MAE comparison between Koopman models with and without the EMG input for all patients.}
        \label{fig:mean absolute error_comparison_patientss}
    \end{subfigure}
    \caption{The prediction evaluation of the Koopman models for the healthy participants and patients.}
    \label{fig:model_predictive performance_comparisons}
\end{figure}

The prediction performance of the two models on healthy participants is shown in Fig.~\ref{fig:emg_loss_comparison_healthy and patientss}.
In particular, incorporating EMG signals consistently reduces the average multi-step prediction error across all prediction horizons. This improvement indicates that neuromuscular activation provides information relevant to state evolution that cannot be inferred from joint kinematics alone. By explicitly taking muscle activation into account, the Koopman framework captures the human--robot coupled dynamics for healthy subjects more accurately.

\subsubsection{Evaluation on the Stroke Patients}

The prediction loss for the stroke patients is shown in Fig.~\ref{fig:emg_loss_comparison_healthy and patientss}. Compared to healthy participants, the overall prediction loss is higher, and it reflects the increased variability and complexity in patient-specific dynamics. To quantify the predictive accuracy, the $R^2$ across all patients is calculated over the entire prediction horizon. The results show that the model incorporating EMG achieves a higher $R^2$ of 0.9962 compared to 0.9946 for the model without EMG. Moreover, as illustrated in Fig.~\ref{fig:mean absolute error_comparison_patientss}, Koopman models that use EMG signals as inputs demonstrate smaller MAE for all patients compared to the model without EMG. 

\subsection{Personalized Assistance Strategy Evaluation}

\begin{table*}[!htpb]
    \centering
    \caption{The tracking RMSE errors in degrees ($^\circ$), compared between personalized and non-personalized models under different frequencies and angle ranges in passive mode for healthy participants P1 to P5.}
    \label{tab:tracking_performance_healthy}
    \begin{tabular*}{\textwidth}{@{\extracolsep{\fill}} l l c c c c c c}
        \hline
        \textbf{Ref. Traj. Freq. (Hz)} & 
        \textbf{Ref. Traj. Peak Knee Motion Range ($^\circ$)} & 
        \textbf{Model Used} & 
        \textbf{P1} & \textbf{P2} & \textbf{P3} & \textbf{P4} & \textbf{P5} \\
        \hline
        0.027 & 90--120 & Pers. & \textbf{2.2933} & \textbf{1.8421} & \textbf{2.0262} & \textbf{0.7189} & \textbf{1.0640} \\
              &          & Non-Pers. & 2.8923 & 2.0143 & 2.4298 & 1.3553 & 2.1641 \\
        \midrule
        0.032 & 90--120 & Pers. & \textbf{3.2650} & \textbf{1.5989} & \textbf{2.2230} & \textbf{0.9069} & \textbf{1.2427} \\
              &          & Non-Pers. & 3.9000 & 2.4224 & 2.7117 & 1.1039 & 2.0770 \\
        \midrule
        0.04  & 90--120 & Pers. & \textbf{4.4369} & \textbf{3.1267} & \textbf{2.7982} & \textbf{1.2030} & \textbf{3.0558} \\
              &          & Non-Pers. & 9.4003 & 4.1941 & 3.1707 & 1.9197 & 3.1998 \\
        \midrule
        0.032 & 90--125 & Pers. & 5.5482 & \textbf{2.5690} & \textbf{3.3543} & \textbf{1.3098} & \textbf{2.2640} \\
              &          & Non-Pers. & 5.3608 & 3.0144 & 3.9561 & 1.8452 & 3.3321 \\
        \midrule
        0.032 & 90--130 & Pers. & \textbf{7.1732} & \textbf{3.6556} & \textbf{5.6207} & 2.2093 & \textbf{2.1623} \\
              &          & Non-Pers. & 7.7117 & 3.7474 & 6.3152 & \textbf{2.1582} & 4.2665 \\
        \hline
    \end{tabular*}
\end{table*}

\begin{table*}[!htpb]
    \centering
    \caption{Comparison of tracking RMSE in degrees ($^\circ$) between personalized and non-personalized models for stroke patients.}
    \label{tab:personalization_patient_horizontal}
    \renewcommand{\arraystretch}{1.3} 
    \setlength{\tabcolsep}{3pt} 
    \begin{tabular*}{\textwidth}{@{\extracolsep{\fill}} l l | c c c c c c c c c c c }
        \hline
        \multicolumn{2}{c|}{\textbf{Mode \& Model}} & \textbf{S1} & \textbf{S2} & \textbf{S3} & \textbf{S4} & \textbf{S5} & \textbf{S6} & \textbf{S7} & \textbf{S8} & \textbf{S9} & \textbf{S10} & \textbf{S11} \\
        \hline
        \multirow{2}{*}{\textbf{Active}} & Pers & 10.78 & \textbf{3.59} & \textbf{7.11} & \textbf{2.55} & \textbf{11.30} & 7.35 & \textbf{7.60} & \textbf{6.56} & \textbf{4.84} & \textbf{4.96} & \textbf{6.72} \\
        & Non-Pers & \textbf{10.13} & 4.73 & 8.71 & 5.21 & 13.81 & \textbf{7.03} & 8.47 & 7.92 & 5.76 & 5.12 & 9.01 \\
        \hline
        \multirow{2}{*}{\textbf{Passive}} & Pers & \textbf{12.95} & \textbf{11.84} & \textbf{10.30} & 11.05 & \textbf{13.57} & \textbf{14.53} & \textbf{11.82} & \textbf{7.17} & \textbf{5.19} & \textbf{5.05} & \textbf{9.42} \\
        & Non-Pers & 13.84 & 13.69 & 13.23 & \textbf{10.68} & 16.09 & 17.59 & 13.01 & 9.83 & 5.90 & 5.27 & 9.77 \\
        \hline
    \end{tabular*}
\end{table*}
Rehabilitation training is inherently subject-specific, as users differ substantially in biomechanics and in their human--robot coupled dynamics. Since the Koopman model in our framework is trained from each participant's own data, the resulting controller is expected to better match each individual's characteristics during knee rehabilitation training. We evaluate the following two modeling strategies for each healthy participant and patient under both active and passive rehabilitation modes:
\begin{itemize}
\item \textbf{Non-personalized model:} The model is trained using the data collected from a different participant.
\item \textbf{Personalized model:} The model is trained with the participant's own data.
\end{itemize}

\subsubsection{Evaluation Among the Healthy Participants}
\begin{figure}[!htpb]
\centering
\includegraphics[width=1\columnwidth]{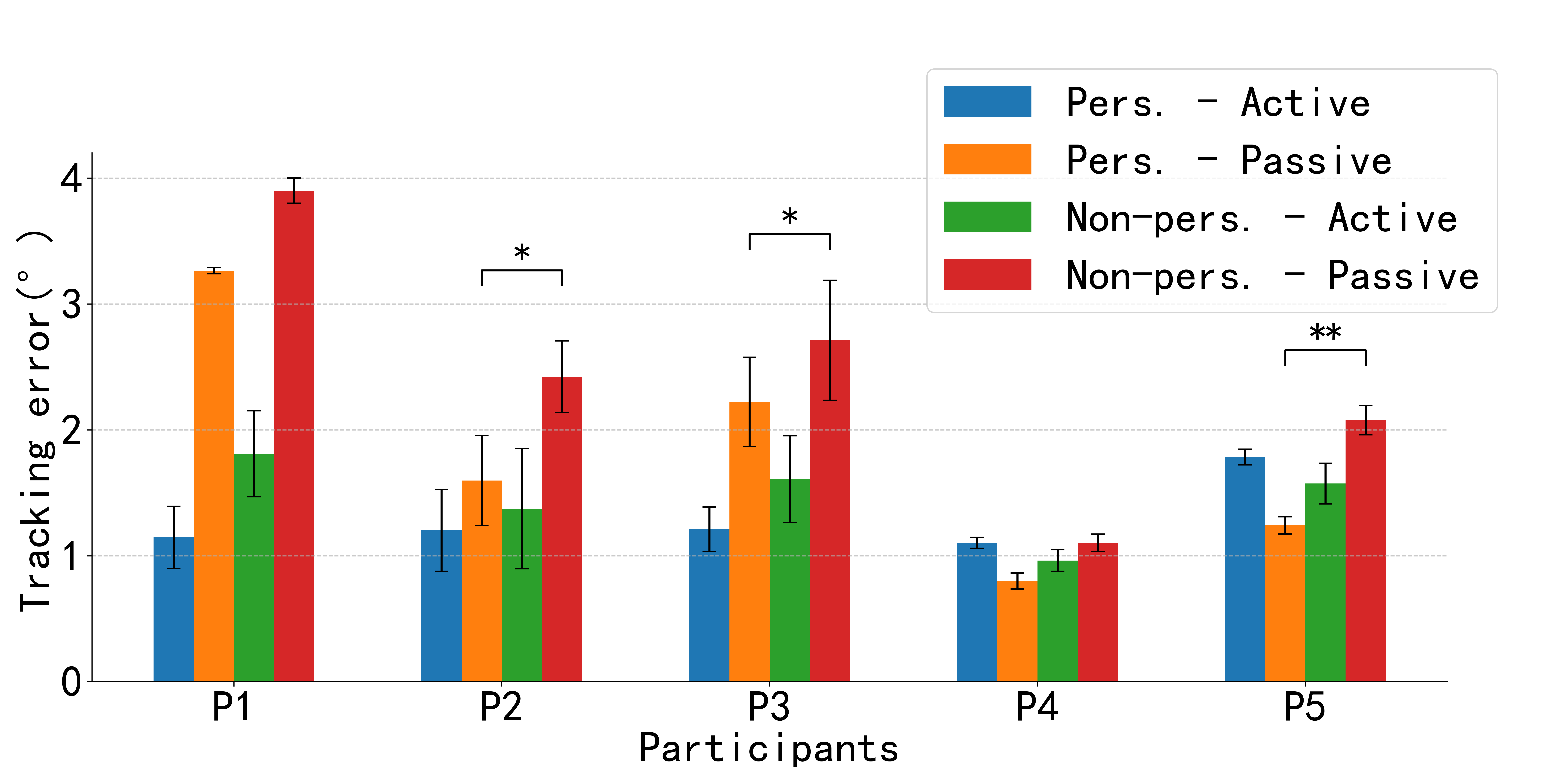}
\caption{The average tracking errors of personalized and non-personalized human-robot coupled dynamics in both active and passive modes among the healthy participants.}
\label{fig:personalization_tracking_error}
\end{figure}
As illustrated in Fig.~\ref{fig:personalization_tracking_error}, the benefit of personalization is particularly pronounced in passive rehabilitation training. The markers ``*'' and ``**'' indicate statistically significant differences between the personalized and non-personalized models under the same rehabilitation mode (``$*$'' indicates $p < 0.05$ and ``$**$'' indicates $p < 0.01$). For example, for participant P5 in passive mode, the KMPC tracking error under the non-personalized model is approximately $2.1^\circ$ on average, whereas the personalized model reduces the error to $1.2^\circ$. Similar statistically significant reductions are also observed for participants P2 and P3, with the passive-mode tracking errors decreasing from $2.4^\circ$ to $1.6^\circ$ (P2) and from $2.7^\circ$ to $2.2^\circ$ (P3), respectively. 


On the other hand, since the human--robot coupled dynamics are identified from the data collected while the participants track a prescribed reference trajectory (Section~\ref{sec:experimental_procedure}), we further evaluate the generalizability of the personalized controller to reference trajectories with different frequencies and peak knee motion ranges. More specifically, the healthy participants perform passive rehabilitation training with different sinusoidal reference trajectories, where their knee is driven by the soft exoskeleton controlled by the KMPC controller using either personalized or non-personalized human--robot coupled models. As shown in Tab.~\ref{tab:tracking_performance_healthy}, when the reference frequency varies (0.027~Hz, 0.032~Hz, and 0.04~Hz) with a fixed peak range ($90^\circ$--$120^\circ$), KMPC with a personalized model achieves the lower RMSE across all healthy participants. Similarly, when the peak range increases (from $90^\circ$--$120^\circ$ to $90^\circ$--$130^\circ$) with a fixed frequency, KMPC with a personalized model continues to outperform the non-personalized model for most of the healthy participants. These results support generalization within the tested seated knee-rehabilitation operating region; broader trajectory-generalization limitations are discussed in Sec.~V.

\subsubsection{Evaluation Among the Patients}
For patient experiments, a non-personalized baseline model is constructed by pre-training a Koopman model using data collected from a healthy participant (first author: 80~kg, 183~cm, age 24). This cross-subject reference contains no data from the target patient, but its healthy source makes health status an additional variable, so the observed performance differences should not be attributed solely to personalization. Due to the limited daily tolerance of stroke patients, the patient experiments focus on the standard rehabilitation trajectory rather than multiple additional sinusoidal references.

As shown in Tab.~\ref{tab:personalization_patient_horizontal}, in passive rehabilitation mode, KMPC with a personalized model achieves the lowest RMSE for 10 out of 11 patients. For S4, the personalized model achieves a tracking RMSE comparable to the non-personalized model. For most patients in the tested cohort, KMPC using a patient-specific model therefore achieves lower tracking errors than this particular cross-subject reference.

A similar trend is observed in active rehabilitation mode: the personalized model outperforms the non-personalized baseline for 9 out of 11 patients. For S1 and S6, although the personalized model does not yield the lowest RMSE, its performance still remains comparable to the non-personalized model.


\subsection{Comparisons with the Controller Benchmarks}

\begin{table*}[!htpb]
\centering
\caption{Comparison of tracking RMSE in degrees ($^\circ$) between KMPC and controller baselines for healthy participants (P1--P5) and stroke patients (S1--S11). 
For participants S8--S11, the comparison with ILC is included.}
\label{tab:kmpc_pid_ilc_horizontal}
\renewcommand{\arraystretch}{1.25}
\setlength{\tabcolsep}{3pt}

\begin{tabular*}{\textwidth}{@{\extracolsep{\fill}} l l | c c c c c c c c c c c c c c c c}
\hline
\multicolumn{2}{c|}{\textbf{Mode \& Model}} 
& \textbf{P1} & \textbf{P2} & \textbf{P3} & \textbf{P4} & \textbf{P5}
& \textbf{S1} & \textbf{S2} & \textbf{S3} & \textbf{S4} & \textbf{S5} & \textbf{S6} & \textbf{S7}
& \textbf{S8} & \textbf{S9} & \textbf{S10} & \textbf{S11} \\
\hline

\multirow{3}{*}{\textbf{Active}} 
& \textbf{KMPC} 
& \textbf{1.147} & \textbf{1.203} & \textbf{1.211} & \textbf{1.103} & 1.785
& \textbf{10.781} & \textbf{3.585} & \textbf{7.114} & \textbf{2.545} & \textbf{11.296} & \textbf{7.349} & \textbf{7.604}
& \textbf{6.559} & 4.837 & \textbf{4.958} & \textbf{6.715} \\

& \textbf{PID} 
& 1.410 & 1.360 & 1.588 & 1.705 & \textbf{1.639}
& 10.830 & 5.945 & 7.389 & 6.608 & 13.863 & 7.976 & 10.776
& 7.428 & 6.103 & 7.953 & 9.567 \\

& \textbf{ILC} 
& -- & -- & -- & -- & --
& -- & -- & -- & -- & -- & -- & --
& 6.861 & \textbf{4.444} & 6.144 & 7.288 \\

\hline

\multirow{3}{*}{\textbf{Passive}} 
& \textbf{KMPC} 
& \textbf{3.265} & \textbf{1.599} & \textbf{2.223} & \textbf{0.907} & \textbf{1.243}
& \textbf{12.950} & \textbf{11.839} & \textbf{10.298} & \textbf{11.054} & \textbf{13.568} & 14.529 & \textbf{11.818}
& \textbf{7.169} & \textbf{5.188} & \textbf{5.053} & \textbf{9.421} \\

& \textbf{PID} 
& 5.640 & 2.380 & 3.082 & 2.254 & 2.131
& 15.794 & 13.002 & 11.359 & 15.731 & 14.820 & \textbf{14.041} & 12.977
& 8.591 & 6.918 & 6.642 & 12.907 \\

& \textbf{ILC} 
& -- & -- & -- & -- & --
& -- & -- & -- & -- & -- & -- & --
& 7.890 & 5.381 & 5.055 & 9.929 \\

\hline
\end{tabular*}
\end{table*}

The tracking performance of the proposed KMPC is compared against the conventional PID controller across all healthy participants (P1--P5) and stroke patients (S1--S11).
The tracking performance is further compared with ILC for the four stroke patients (S8--S11) for whom ILC data are available.
The comparison results are summarized in Tab.~\ref{tab:kmpc_pid_ilc_horizontal}.

\begin{figure}[!htpb]
\centering
\includegraphics[width=1\columnwidth]{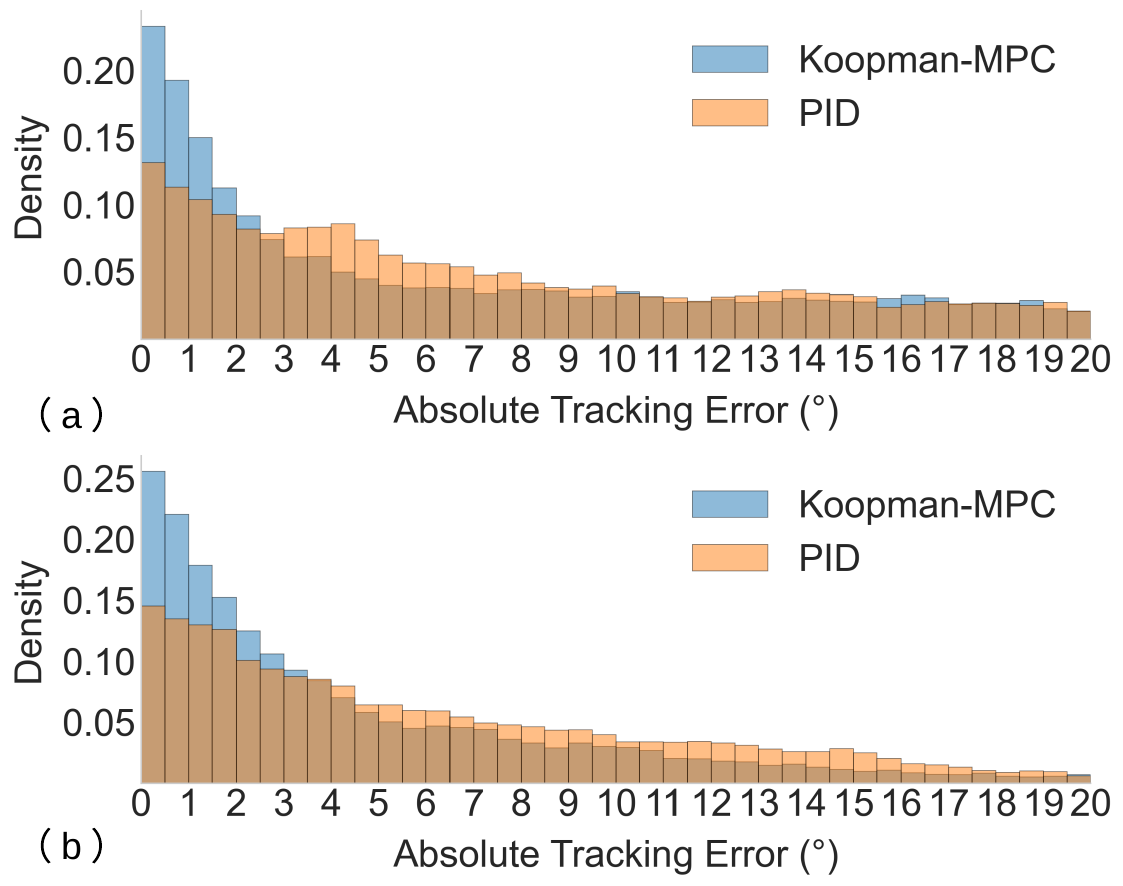}
\caption{The tracking error histograms of the KMPC and PID controllers for both healthy participants and stroke patients. (a) Tracking error distribution in passive mode. (b) Tracking error distribution in active mode.}
\label{fig:kmpc_tracking_modes}
\end{figure}

\subsubsection{Tracking Performance Evaluation}

In active rehabilitation training, KMPC achieves lower RMSE than PID for 4 out of 5 healthy participants and for all stroke patients. For the four patients with ILC comparisons, KMPC achieves lower RMSE than ILC in three out of four active-mode cases. In passive rehabilitation training, KMPC yields lower RMSE than PID for all healthy participants and for 10 out of 11 stroke patients, and it also achieves lower RMSE than ILC for all four patients with ILC comparisons. Overall, these results show that KMPC provides a stronger tracking performance than PID and achieves lower tracking errors than ILC in most of the available ILC comparisons.

To further examine the tracking error distribution, we pool data from both healthy participants and patients and plot the combined tracking error histogram. 
As shown in Fig.~\ref{fig:kmpc_tracking_modes}, KMPC outperforms conventional PID in both passive and active rehabilitation modes. 
In particular, compared to PID, the tracking error distributions under KMPC are more concentrated around zero and exhibit reduced spread, especially in passive rehabilitation training.

\subsubsection{Actuation Effort Evaluation}

\begin{figure}[!htpb]
\centering
\includegraphics[width=1\columnwidth]{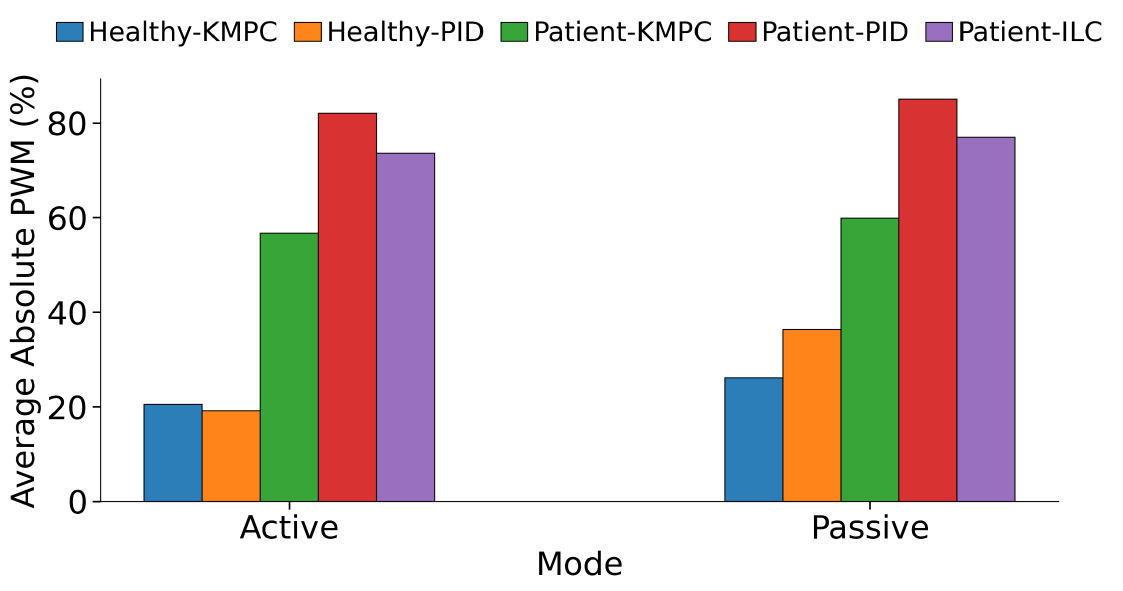}
\caption{Comparison of the average absolute PWM duty-cycle values in both active and passive modes.}
\label{fig:Average_PWM_Comparison}
\end{figure}

As shown in Fig.~\ref{fig:Average_PWM_Comparison}, KMPC generally produces the smallest average absolute PWM wave duty cycle value. For the healthy participants, the average absolute PWM wave duty cycle value under KMPC is slightly higher than PID in active mode, but is substantially lower in passive mode. 
Compared with PID, KMPC generally reduces the average absolute PWM wave duty cycle value. For the stroke patients with ILC comparisons, KMPC also shows lower actuation effort in the tested cases.
These results suggest that the proposed KMPC framework provides smoother and more efficient actuation while maintaining the reference trajectory tracking performance. Overall, the results demonstrate that personalized KMPC enables effective human--robot interaction by providing sufficient assistance while reducing both the user's neuromuscular effort and the exoskeleton's actuation effort.

\subsection{Computation Time Consumption and Assistance Effectiveness}
To further evaluate the practicality of the proposed framework for rehabilitation training, we analyze two aspects: (i) the real-time computational performance of the controller and (ii) the effectiveness of assistance during human--robot interaction.

\subsubsection{Computation Time Consumption}

\begin{figure}[!htpb]
\centering
\includegraphics[width=1\columnwidth]{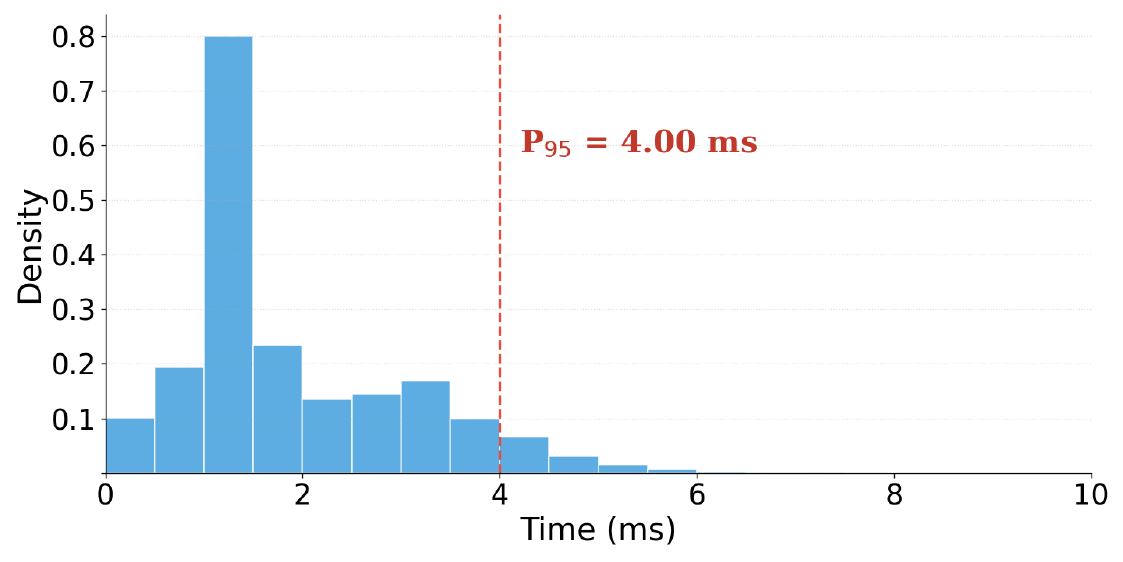}
\caption{The histogram of the computation time of the proposed KMPC framework.}
\label{fig:computation_time}
\end{figure}

Recall that the proposed personalized KMPC is executed on the host computer and the optimized control input is transmitted to the lower-level actuator system for real-time assistance. 
With the implementation method in Sec.~\ref{sec: real_time_implementation}, the controller is able to operate at a frequency of 50 Hz on the host computer, which matches the sensor data acquisition frequency and the Bluetooth communication rate of the rehabilitation platform. The computation time consumption histogram is illustrated in Fig.~\ref{fig:computation_time}.
In particular, the average computation time of each control cycle is approximately 1.6 ms and the 95th-percentile computation time is 4 ms, which are smaller than the 20 ms sampling interval. This demonstrates that the proposed personalized KMPC framework satisfies the real-time requirements.

\subsubsection{Assistance Effectiveness}

To further assess the effectiveness of the proposed framework, we evaluate the reduction in neuromuscular effort during rehabilitation training by comparing the (unnormalized) RMS amplitudes of the rectus femoris EMG signal (EMG1) and the biceps femoris EMG signal (EMG2) under three conditions: (1) no exoskeleton assistance, (2) personalized KMPC-assisted active mode, and (3) personalized KMPC-assisted passive mode. All data in this analysis are collected from the healthy participant P4.

\begin{figure}[!htpb]
\centering
\includegraphics[width=1\columnwidth]{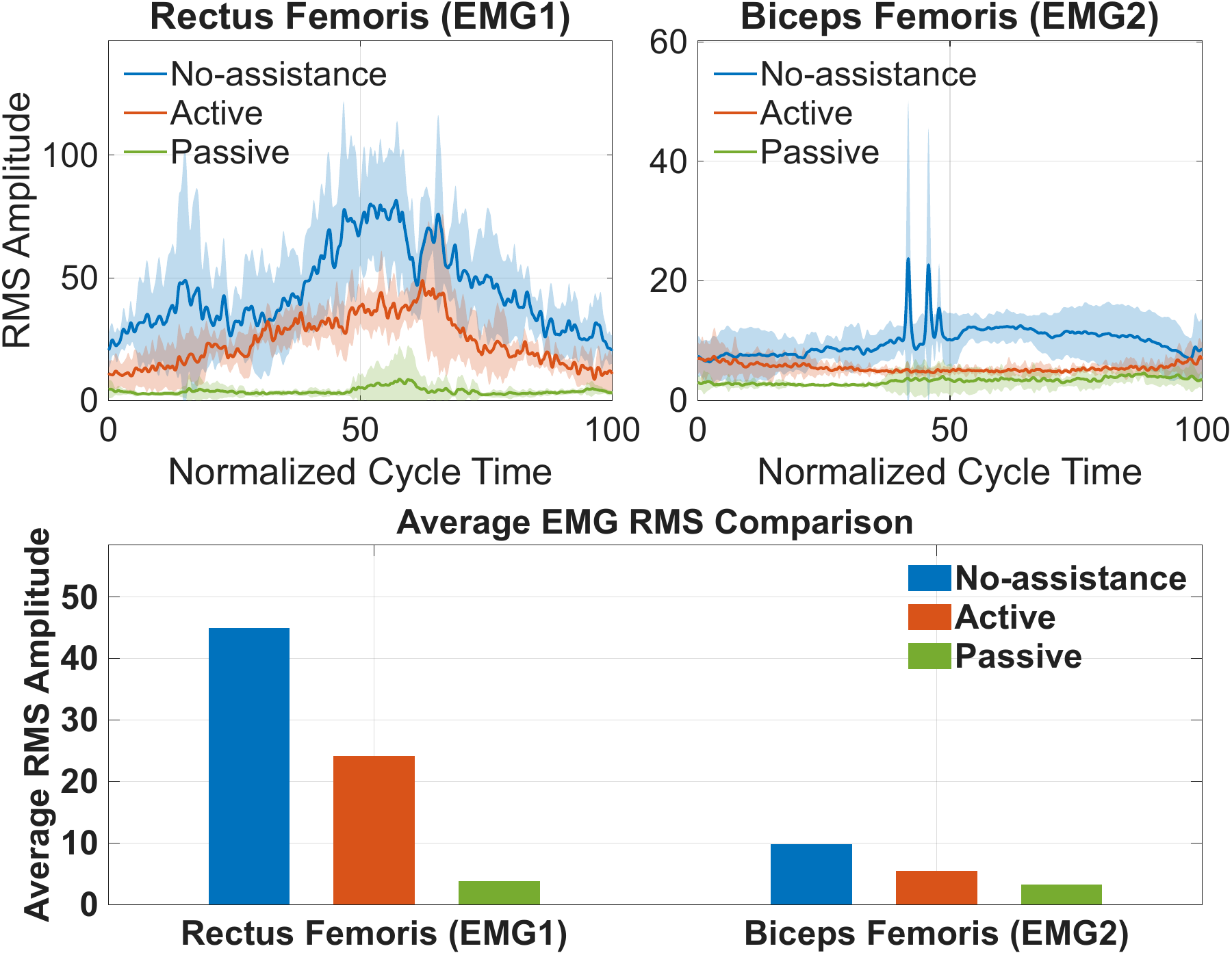}
\caption{Comparison of RMS amplitudes across the following three conditions: (1) no-exoskeleton assistance, (2) KMPC-assisted active mode, and (3) KMPC-assisted passive mode.}
\label{fig:Emg_ifassistance}
\end{figure}

As illustrated in Fig.~\ref{fig:Emg_ifassistance}, the no-assistance condition exhibits the highest EMG activation levels throughout the cycle. With soft exoskeleton assistance, the EMG activation levels decrease substantially in both active and passive rehabilitation modes. As expected, passive mode yields the lowest EMG levels, since the soft exoskeleton provides the majority of the required torque and the participant is instructed to remain relaxed.

These results indicate that, in the evaluated healthy-participant case, the proposed personalized KMPC framework reduces muscular effort during assisted rehabilitation training while maintaining stable trajectory tracking.
This shows that the controller can provide sufficient assistance and improve the overall human--robot interaction performance.

\section{Discussion and Limitations}
This study develops and validates an EMG-informed Koopman modeling and KMPC control framework for seated soft knee rehabilitation. The experiments with healthy participants and eleven stroke patients demonstrate the feasibility of the proposed framework and its subject-specific tracking performance in the tested rehabilitation setting. Taken together, these results clarify the respective roles of EMG information, subject-specific modeling, and operating-region coverage in the proposed framework, while providing a basis for broader validation.

The comparison in Fig.~\ref{fig:model_predictive performance_comparisons} shows that adding EMG consistently improves model prediction, although the magnitude of the improvement is modest. This magnitude can be partly explained by the current rehabilitation task and excitation protocol. Since the seated knee trajectories are slow and repetitive, the recent knee-angle history already provides strong predictive information. In addition, the patient-safe protocol avoids aggressive or randomized excitation, which limits voluntary-activation variability. Even so, the consistent improvement obtained by adding EMG indicates that neuromuscular activation remains a useful complementary input for the Koopman model. EMG may become more beneficial with stronger voluntary participation, especially during movement initiation or direction-transition phases in tracking training. Its practical value may also appear in assistance timing, interaction comfort, and user-perceived assistance, which will be examined in future work through richer but safety-constrained task variations and interaction measurements.

Temporal alignment is central to incorporating EMG into the learned human--robot dynamics.
The fixed 200-ms electromechanical delay (EMD) is a practical alignment choice rather than a subject-specific estimate. With the 20~ms sampling period, it corresponds to a 10-sample shift that can be implemented directly in the real-time controller. Since EMD can vary across muscles, contraction conditions, and individuals, this fixed value may not provide the best temporal alignment for every participant. Nevertheless, the prediction performance comparison between Koopman models with and without EMG shows that the delayed EMG still contributes useful information in the tested cases. Subject-specific EMD estimation may further improve temporal alignment in future work.

In the control formulation, delayed EMG is used as a feedforward input representing the participant's active neuromuscular contribution in the predicted human--robot coupled dynamics. The KMPC optimizes robot actuation while accounting for this contribution, but it does not explicitly infer the participant's intended trajectory or estimate a required assistance level, nor does it use a prescribed effort-to-assistance mapping. We therefore describe the resulting control as EMG-informed assistance.

The personalization results further demonstrate the value of matching the learned dynamics to the individual and to the operating region covered by the identification data. When the non-personalized model already approximates a participant reasonably well, the room for statistically significant improvement becomes smaller; nevertheless, the personalized model still yields lower tracking errors for most participants. Accordingly, the patient comparison is interpreted as patient-specific identification versus transfer of a model from another user; population-level or leave-one-patient-out references constructed from other stroke patients will be considered in future work. The identification data collected under closed-loop PID tracking provide task-relevant excitation. Although this protocol does not establish a formal persistent-excitation guarantee, the current experimental evidence for model validity is limited to the intended seated knee-rehabilitation operating region. The additional tests with sinusoidal reference trajectories of different frequencies and amplitudes extend this evidence by showing that the learned Koopman model generalizes reasonably within the tested operating region. Future validation with non-periodic and other task-oriented lower-limb movements will examine how this generalization extends beyond the present class of sinusoidal trajectories.

In the experiments, PID serves as a conventional baseline, while ILC provides a meaningful learning-based benchmark for repetitive rehabilitation tasks. The lower RMSE achieved by KMPC in most comparisons supports the effectiveness of the personalized predictive formulation in the evaluated rehabilitation setting. Comparisons with other data-driven predictive controllers can further clarify its relative performance in future studies.

The patient experiments extend these findings to a clinically relevant cohort with inter-participant variability. All recruited patients are within Brunnstr\"om stages 3--5 and retain residual seated knee-movement ability. No EMG-based screening or signal-quality preselection is performed. The results therefore suggest that the subject-specific Koopman models can accommodate the range of inter-patient EMG variations observed in the recruited cohort. However, because EMG quality is not formally assessed, the present study does not establish a minimum EMG activation or signal-quality requirement. Applicability to patients with more severe impairment, absent muscle activation, or unusable EMG signals remains to be evaluated. The present short-term experiments establish modeling and control feasibility; longer intervention studies can further assess user adaptation, comfort evolution, and rehabilitation outcomes.

For clinical translation, the modeling and control results should be considered together with hardware-level usability. Interface comfort, contact pressure distribution, and relative slippage are important factors for wearable rehabilitation systems. Since the primary focus of this work is the EMG-informed Koopman modeling and KMPC control framework, the present hardware evaluation focuses on actuator integration, sensing, safety design, and experimental feasibility of the soft knee exoskeleton. The connector and strap structures are nevertheless designed with wearability considerations in mind, including improved force transmission, fixation stability, and reduced relative slippage.

The acoustic measurements provide an additional practical characterization of prototype usability. The prototype acoustic noise is measured in a single 30-second trial using a Class~2 AWA5636 sound level meter positioned 1~m from the pneumatic drive unit, with A-frequency weighting and Fast time weighting. The background noise level is 42.3~dBA; the average displayed sound level is 67~dBA during rehabilitation operation, and the maximum displayed sound level is 71.4~dBA. In line with the personal-care robot considerations emphasized by ISO 13482, future prototype development should further consider safety, user comfort, acceptability, and hardware robustness. These considerations include not only acoustic comfort, but also fail-safe operation, stable fixation, and comfortable human--device interaction during repeated clinical use. Hardware optimization will focus on reducing transmitted pump vibration and pneumatic noise through vibration-isolated mounting and acoustically treated structural enclosures.

\section{Conclusion}
This paper presents a soft knee rehabilitation exoskeleton based on an origami-inspired pneumatic actuator, together with a data-driven control framework for personalized assistance. The developed exoskeleton is lightweight and easy to wear, and its connector and fixation structures are designed to support safe, compliant, and effective force transmission during seated knee rehabilitation. By modeling the human and the soft exoskeleton as a unified system, a Deep Koopman Network is employed to learn a linear representation of the human--robot coupled dynamics from measured knee joint angles, EMG signals, and actuation commands. Based on the learned model, a KMPC framework is developed to provide personalized assistance during rehabilitation training.

Experimental results from both healthy participants and stroke patients demonstrate that the proposed Deep Koopman Network captures human--robot interaction dynamics and that incorporating EMG signals improves prediction accuracy. Owing to its data-driven and subject-specific nature, the learned model adapts effectively to individual differences in biomechanics and neuromuscular activation. The comparison with PID across both healthy participants and stroke patients, together with the additional ILC benchmark on representative stroke patients, supports the tracking effectiveness of the proposed KMPC framework in both passive and active rehabilitation modes. Furthermore, the results suggest that the framework can reduce actuation effort and, in the evaluated assistance experiments, help reduce users' neuromuscular effort.

\ifCLASSOPTIONcaptionsoff
  \newpage
\fi
\bibliographystyle{unsrt}
\bibliography{references}

\end{document}